\documentclass{article}

\usepackage[preprint]{neurips_2021}

\usepackage[utf8]{inputenc} 
\usepackage[T1]{fontenc}    
\usepackage{hyperref}       
\usepackage{url}            
\usepackage{booktabs}       
\usepackage{amsfonts}       
\usepackage{nicefrac}       
\usepackage{microtype}      
\usepackage{xcolor}         

\usepackage{amsmath,amsfonts,bm}

\def\eqref#1{equation~\ref{#1}}

\def\1{\bm{1}}

\def\va{{\bm{a}}}

\def\vc{{\bm{c}}}

\def\vn{{\bm{n}}}

\def\vx{{\bm{x}}}

\DeclareMathAlphabet{\mathsfit}{\encodingdefault}{\sfdefault}{m}{sl}
\SetMathAlphabet{\mathsfit}{bold}{\encodingdefault}{\sfdefault}{bx}{n}

\def\sD{{\mathbb{D}}}

\newcommand{\R}{\mathbb{R}}

\usepackage{graphicx}
\usepackage{subcaption}
\usepackage{booktabs} 
\usepackage{enumitem}

\title{Scalable Rule-Based Representation Learning for\\ Interpretable Classification}

\author{Zhuo Wang\textsuperscript{\rm 1}, Wei Zhang\textsuperscript{\rm 2}\thanks{Corresponding authors.} , Ning Liu\textsuperscript{\rm 1}, Jianyong Wang\textsuperscript{\rm 1}\footnotemark[1]\\ 
\textsuperscript{\rm 1}Department of Computer Science and Technology, Tsinghua University\\
\textsuperscript{\rm 2}School of Computer Science and Technology, Shanghai Key Laboratory of \\Trustworthy Computing, East China Normal University\\

wang-z18@mails.tsinghua.edu.cn, \{zhangwei.thu2011, victorliucs\}@gmail.com,\\ jianyong@tsinghua.edu.cn 
}

\begin{document}

\maketitle

\begin{abstract}
  Rule-based models, e.g., decision trees, are widely used in scenarios demanding high model interpretability for their transparent inner structures and good model expressivity. However, rule-based models are hard to optimize, especially on large data sets, due to their discrete parameters and structures. Ensemble methods and fuzzy/soft rules are commonly used to improve performance, but they sacrifice the model interpretability. To obtain both good scalability and interpretability, we propose a new classifier, named Rule-based Representation Learner (RRL), that automatically learns interpretable non-fuzzy rules for data representation and classification. To train the non-differentiable RRL effectively, we project it to a continuous space and propose a novel training method, called Gradient Grafting, that can directly optimize the discrete model using gradient descent. An improved design of logical activation functions is also devised to increase the scalability of RRL and enable it to discretize the continuous features end-to-end. Exhaustive experiments on nine small and four large data sets show that RRL outperforms the competitive interpretable approaches and can be easily adjusted to obtain a trade-off between classification accuracy and model complexity for different scenarios. Our code is available at : \url{https://github.com/12wang3/rrl}.
\end{abstract}

\section{Introduction}
Although Deep Neural Networks (DNNs) have achieved impressive results in various machine learning tasks \citep{goodfellow2016deep}, rule-based models, benefiting from their transparent inner structures and good model expressivity, still play an important role in domains demanding high model interpretability, such as medicine, finance, and politics \citep{doshi2017towards}.
In practice, rule-based models can easily provide explanations for users to earn their trust and help protect their rights \citep{molnar2019,lipton2016mythos}. By analyzing the learned rules, practitioners can understand the decision mechanism of models and use their knowledge to improve or debug the models \citep{chu2018exact}.
Moreover, even if post-hoc methods can provide interpretations for DNNs, the interpretations from rule-based models are more faithful and specific \citep{murdoch2019interpretable}.
However, conventional rule-based models are hard to optimize, especially on large data sets, due to their discrete parameters and structures, which limit their application scope. 
To take advantage of rule-based models in more scenarios, we urgently need to answer such a question: \textit{how to improve the scalability of rule-based models while keeping their interpretability?}

Studies in recent years provide some solutions to improve conventional rule-based models in different aspects. Ensemble methods and soft/fuzzy rules are proposed to improve the performance and scalability of rule-based models but at the cost of model interpretability \citep{ke2017lightgbm,breiman2001random,irsoy2012soft}. Bayesian frameworks are also leveraged to more reasonably restrict and adjust the structures of rule-based models \citep{letham2015interpretable,wang2017bayesian,yang2017scalable}. However, due to the non-differentiable model structure, they have to use methods like MCMC or Simulated Annealing, which could be time-consuming for large models. 
Another way to improve rule-based models is to let a high-performance but complex model (e.g., DNN) teach a rule-based model \citep{frosst2017distilling,ribeiro2016should}. However, learning from a complex model requires soft rules, or the fidelity of the student model is not guaranteed. 
The recent study~\cite{wang2020transparent} tries to extract hierarchical rule sets from a tailored neural network. 
When the network is large, the extracted rules could behave quite differently from the neural network and become useless in most cases. 
Nevertheless, combined with binarized networks \citep{cour2015bconnect}, it inspires us that we can search for the discrete solutions of interpretable rule-based models in a continuous space leveraging effective optimization methods like gradient descent.

In this paper, we propose a novel rule-based model named \textbf{Rule-based Representation Learner (RRL)} (see Figure~\ref{fig:RRL}). We summarize the key contributions as follows:

\begin{itemize}[leftmargin=*,itemsep=0pt,parsep=0.0em,topsep=0.0em,partopsep=0.0em]
\item To achieve good model \textbf{transparency and expressivity}, RRL is formulated as a hierarchical model, with layers supporting automatic feature discretization, rule-based representation learning in flexible conjunctive and disjunctive normal forms, and rule importance evaluation.

\item To facilitate \textbf{training effectiveness}, RRL exploits a novel gradient-based discrete model training method, Gradient Grafting, that directly optimizes the discrete model and uses the gradient information at both continuous and discrete parametric points to accommodate more scenarios.

\item To ensure \textbf{data scalability}, RRL utilizes improved logical activation functions to handle high-dimensional features. By further combining the improved logical activation functions with a tailored feature binarization layer, it realizes the continuous feature discretization end-to-end.

\item We conduct experiments on nine small data sets and four large data sets to validate the advantages, i.e., good \textbf{accuracy and interpretability}, of our model over other representative classification models. The benefits of the model's key components are also verified by the experiments.
\end{itemize}

\section{Related Work}
\textbf{Rule-based Models}. Decision tree, rule list, and rule set are the widely used structures in rule-based models. For their discrete parameters and non-differentiable structures, we have to train them by employing various heuristic methods \citep{Quinlan:1993:CPM:583200,breiman2017classification,cohen1995fast}, which may not find the globally best solution or a solution with close performance. Alternatively, train them with search algorithms \citep{wang2017bayesian,angelino2017learning}, which could take too much time on large data sets. 
In recent studies, Bayesian frameworks are leveraged to restrict and adjust model structure more reasonably \citep{letham2015interpretable,wang2017bayesian,yang2017scalable}.
\cite{lakkaraju2016interpretable} learns independent if-then rules with smooth local search. 
However, except for heuristic methods, most existing rule-based models need frequent itemsets mining and/or long-time searching, which limits their applications. Moreover, it is hard for these rule-based models to get comparable performance with complex models like Random Forest.

Ensemble models like Random Forest \citep{breiman2001random} and Gradient Boosted Decision Trees \citep{chen2016xgboost,ke2017lightgbm} have better performance than the single rule-based model. However, since the decision is made by hundreds of models, ensemble models are commonly not considered as interpretable models \citep{hara2016making}. Soft or fuzzy rules are also used to improve model performance \citep{irsoy2012soft,ishibuchi2005rule}, but non-discrete rules are much harder to understand than discrete ones. 
Deep Neural Decision Tree \citep{yang2018deep} is a tree model realized by neural networks with the help of soft binning function and Kronecker product. However, due to the use of Kronecker product, it is not scalable with respect to the number of features.
Other studies try to teach the rule-based model by a complex model, e.g., DNN, or extract rule-based models from complex models \citep{frosst2017distilling,ribeiro2016should,wang2020transparent}. However, the fidelity of the student model or extracted model is not guaranteed. 

\textbf{Gradient-based Discrete Model Training}. The gradient-based discrete model training methods are mainly proposed to train binary or quantized neural networks for network compression and acceleration.
\cite{cour2015bconnect,courbariaux2016binarized} propose to use the Straight-Through Estimator (STE) for binary neural network training. However, STE requires gradient information at discrete points, which limits its applications. 
ProxQuant \citep{bai2018proxquant} formulates quantized network training as a regularized learning problem and optimizes it via the prox-gradient method. ProxQuant can use gradients at non-discrete points but cannot directly optimize for the discrete model.
The RB method \citep{wang2020transparent} trains a neural network with random binarization for its weights to ensure the discrete and the continuous model behave similarly.
However, when the model is large, the differences between the discrete and the continuous model are inevitable. Gumbel-Softmax estimator \citep{jang2016categorical} generates a categorical distribution with a differentiable sample. 
However, it can hardly deal with a large number of variables, e.g., the weights of binary networks, for it is a biased estimator. 
Our method, i.e., Gradient Grafting, is different from all the aforementioned works in using gradient information of both discrete and continuous models in each backpropagation.
\section{Rule-based Representation Learner}
\noindent \textbf{Notation Description.}
Let $\sD=\{(X_1,Y_1),\dots,(X_N,Y_N)\}$ denote the training data set with $N$ instances, where $X_i$ is the observed feature vector of the $i$-th instance with the $j$-th entry as $X_{i,j}$, and $Y_i$ is the corresponding one-hot class label vector, $i\in\{1, \dots, N\}$.
Each feature value can be either discrete or continuous.
All the classes take discrete values, and the number of class labels is denoted by $M$.
We use one-hot encoding to represent all discrete features as binary features.
Let $C_i\in \R^m$ and $B_i\in \{0,1\}^b$ denote the continuous feature vector and the binary feature vector of the $i$-th instance respectively.
Therefore, $X_i=C_i\oplus B_i$, where $\oplus$ represents the operator that concatenates two vectors.
Throughout this paper, we use 1 (True) and 0 (False) to represent the two states of a Boolean variable. Thus each dimension of a binary feature vector corresponds to a Boolean variable.
\begin{figure*}
\centering
  \begin{subfigure}[c]{0.56\linewidth}
    \includegraphics[width=\linewidth]{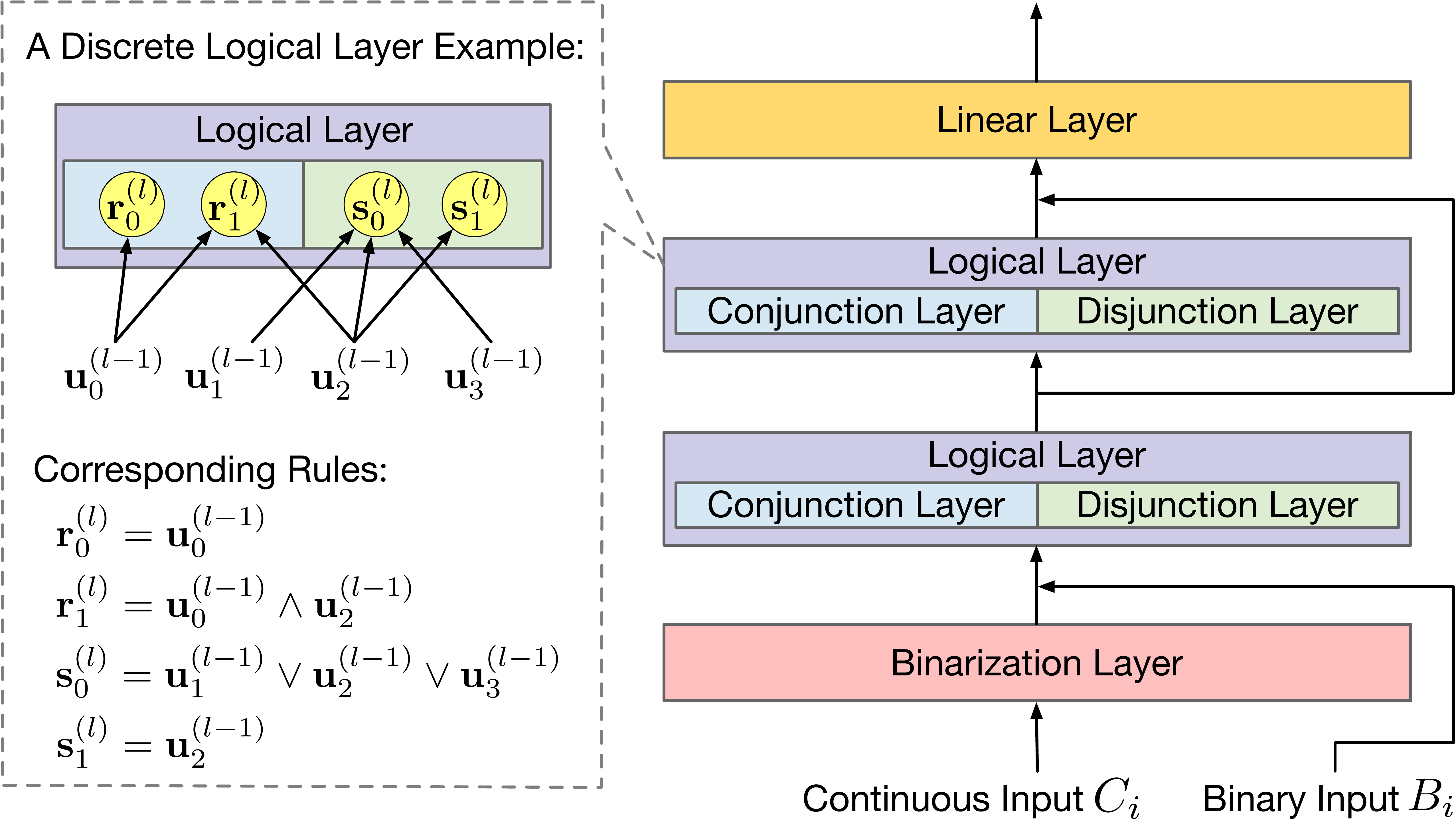}
    \vspace{-10pt}
    \caption{}
    \label{fig:RRL}
  \end{subfigure}
  \hfill
  \begin{subfigure}[c]{0.33\linewidth}
    \includegraphics[width=\linewidth]{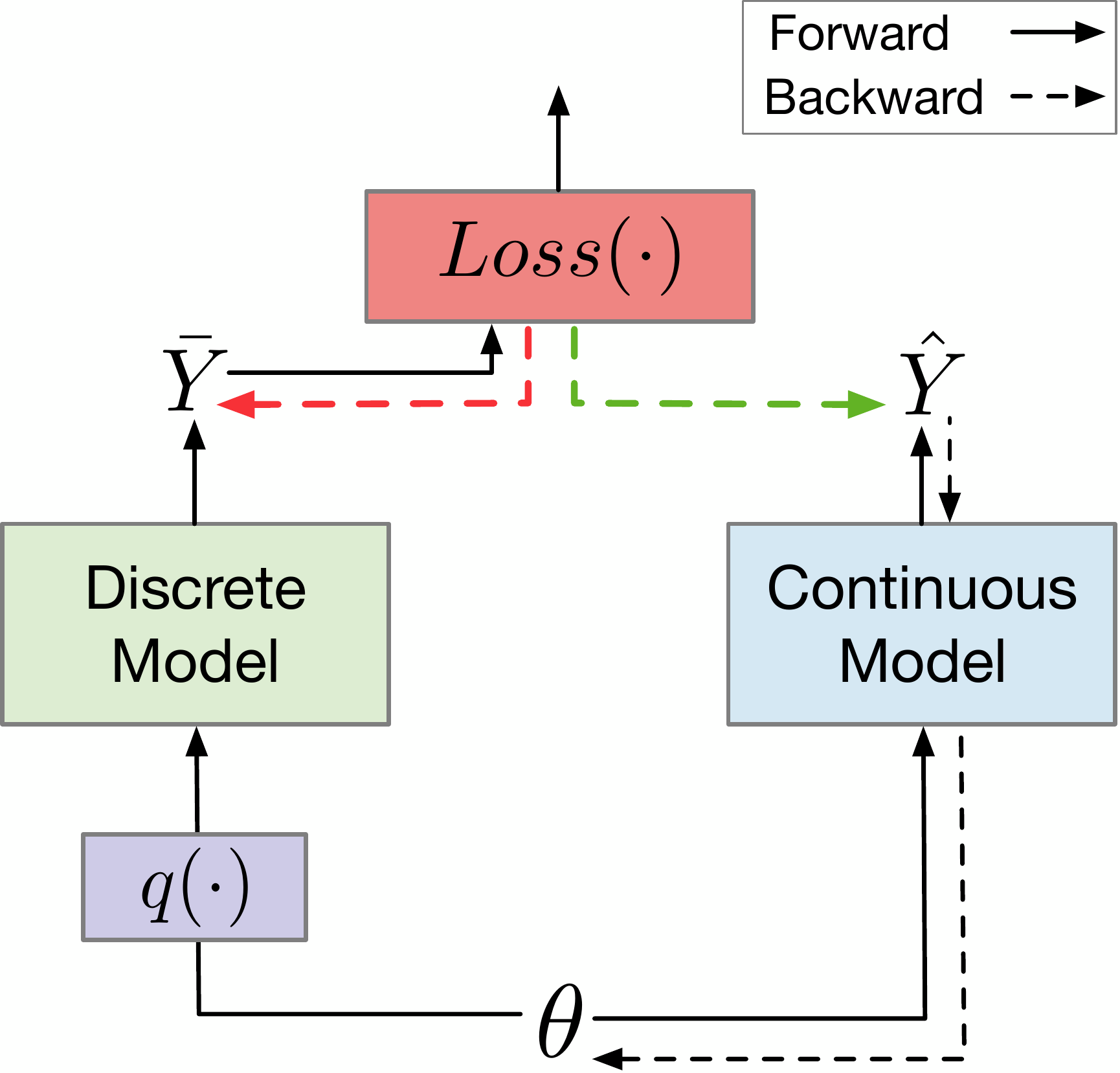}
    \vspace{-10pt}
    \caption{}
    \label{fig:gradient_grafting}
  \end{subfigure}
\vspace{-5pt}
\caption{(a) A Rule-based Representation Learner example. The dashed box shows an example of a discrete logical layer and its corresponding rules. (b) A simplified computation graph of Gradient Grafting. Arrows with solid lines represent forward pass while arrows with dashed lines represent backpropagation. The green arrow denotes the grafted gradient, a copy of the gradient represented by the red arrow. After grafting, there exists a backward path from the loss function to the parameter $\theta$.}
\label{fig:selector}
\end{figure*}

\noindent \textbf{Overall Structure.}
A Rule-based Representation Learner (RRL), denoted by $\mathcal{F}$, is a hierarchical model consisting of three different types of layers. Each layer in RRL not only contains a specific number of nodes but also has trainable edges connected with its previous layer. Let $\mathcal{U}^{(l)}$ denote the $l$-th layer of RRL, $\mathbf{u}^{(l)}_{j}$ indicate the $j$-th node in the layer, and $\vn_l$ represent the corresponding number of nodes, $l\in \{0,\dots,L\}$. The output of the $l$-th layer is a vector containing the values of all the nodes in the layer.
For ease of expression, we denote this vector by $\mathbf{u}^{(l)}$.
There are only one binarization layer, i.e., $\mathcal{U}^{(0)}$,  and one linear layer, i.e., $\mathcal{U}^{(L)}$, in RRL, but the number of middle layers, i.e., logical layers, can be flexibly adjusted according to the specific situation. The logical layers mainly aim to learn the non-linear part of the data, while the linear layer aims to learn the linear part. One example of RRL is shown in Figure \ref{fig:RRL}.

When we input the $i$-th instance to RRL, the binarization layer will first binarize the continuous feature vector $C_i$ into a new binary vector $\bar{C}_i$. Then, $\bar{C}_i$ and $B_i$ are concatenated together as $\mathbf{u}^{(0)}$ and inputted to the first logical layer. 
The logical layers are designed to automatically learn data representations using logical rules, and the stacked logical layers can learn rules in more complex forms. After going through all the logical layers, the output of the last logical layer can be considered as the new feature vector to represent the instance, wherein each feature corresponds to one rule formulated by the original features.
As such, the whole RRL is composed of a feature learner and a linear classifier (linear layer). 
Moreover, the skip connections in RRL can skip unnecessary logical layers.
In what follows, the details of these components will be elaborated.
\subsection{Logical Layer}
\label{section:logical_layer}
Considering the binarization layer needs the help of its following logical layer to binarize features in an end-to-end way, we introduce logical layers first. 
As mentioned above, logical layers can learn data representations using logical rules automatically. To achieve this, logical layers are designed to have a discrete version and a continuous version. The discrete version is used for training, testing and interpretation while the continuous version is only used for training. 
It is worth noting that the discrete RRL indicates the parameter weights of logical layers take discrete values (i.e., 0 or 1) while the parameter weights and biases of the linear layer still take continuous values.

\noindent \textbf{Discrete Version.}
One logical layer consists of one conjunction layer and one disjunction layer. In discrete version, let $\mathcal{R}^{(l)}$ and $\mathcal{S}^{(l)}$ denote the conjunction and disjunction layer of $\mathcal{U}^{(l)}$ ($l\in \{1,2,\dots, L-1\}$) respectively. We denote the $i$-th node in $\mathcal{R}^{(l)}$ by $\mathbf{r}_{i}^{(l)}$, and the $i$-th node in $\mathcal{S}^{(l)}$ by $\mathbf{s}_{i}^{(l)}$.
Specifically speaking, node $\mathbf{r}_{i}^{(l)}$ corresponds to the conjunction of nodes in the previous layer connected with $\mathbf{r}_{i}^{(l)}$, 
while node $\mathbf{s}_{i}^{(l)}$ corresponds to the disjunction of nodes in previous layer connected with $\mathbf{s}_{i}^{(l)}$.
Formally, the two types of nodes are defined as follows:
\begin{equation}
\label{eq:discrete_logic}
\mathbf{r}_{i}^{(l)} =
\bigwedge_{W_{i,j}^{(l,0)}=1}\mathbf{u}_{j}^{(l-1)},\;\;\;\;\;\; \mathbf{s}_{i}^{(l)} = \bigvee_{W_{i,j}^{(l,1)}=1}\mathbf{u}_{j}^{(l-1)},
\end{equation}
where $W^{(l,0)}$ denote the adjacency matrix of the conjunction layer $\mathcal{R}^{(l)}$ and the previous layer $\mathcal{U}^{(l-1)}$, and $W_{i,j}^{(l,0)} \in \{0,1\}$. $W_{i,j}^{(l,0)}=1$ indicates there exists an edge connecting $\mathbf{r}_{i}^{(l)}$ to $\mathbf{u}_{j}^{(l-1)}$, otherwise $W_{i,j}^{(l,0)}=0$. Similarly, $W^{(l,1)}$ is the adjacency matrix of the disjunction layer $\mathcal{S}^{(l)}$ and $\mathcal{U}^{(l-1)}$. Similar to neural networks, we regard these adjacency matrices as the weight matrices of logical layers. $\mathbf{u}^{(l)}=\mathbf{r}^{(l)}\oplus \mathbf{s}^{(l)}$, where $\mathbf{r}^{(l)}$ and $\mathbf{s}^{(l)}$ are the outputs of $\mathcal{R}^{(l)}$ and $\mathcal{S}^{(l)}$ respectively.

The function of the logical layer is similar to the notion ``level'' in \cite{wang2020transparent}. However, one level in that work, which actually consists of two layers, can only represent rules in Disjunctive Normal Form (DNF), while two logical layers in RRL can represent rules in DNF and Conjunctive Normal Form (CNF) at the same time. Connecting nodes in $\mathcal{R}^{(l)}$ with nodes in $\mathcal{S}^{(l-1)}$, we get rules in CNF, while connecting nodes in $\mathcal{S}^{(l)}$ with nodes in $\mathcal{R}^{(l-1)}$, we get rules in DNF. The flexibility of logical layer is quite important. For instance, the length of CNF rule $(\va_1\vee \va_2)\wedge \dots \wedge(\va_{2n-1}\vee \va_{2n})$ is $2n$, but the length of its corresponding DNF rule $(\va_1\wedge \va_3 \dots \wedge \va_{2n-1})\vee \dots \vee(\va_2\wedge \va_4 \dots \wedge \va_{2n})$ is $n\cdot2^{n}$, which means layers that only represent DNF can hardly learn this CNF rule.

\noindent \textbf{Continuous Version.}
Although the discrete logical layers have good interpretability, they are hard to train for their discrete parameters and non-differentiable structures. 
Inspired by the training process of binary neural networks that searches the discrete solution in a continuous space, we extend the discrete logical layer to a continuous version. The continuous version is differentiable, and when we discretize the parameters of a continuous logical layer, we can obtain its corresponding discrete logical layer.
Therefore, the continuous logical layer and its corresponding discrete logical layer can also be considered as sharing the same parameters, but the discrete logical layer needs to discretize the parameters first.

Let $\hat{\mathcal{U}}^{(l)}$ denote the continuous logical layer, and $\hat{\mathcal{R}}^{(l)}$  and $\hat{\mathcal{S}}^{(l)}$ denote the continuous conjunction and disjunction layer respectively, $l\in \{1,2,\dots, L-1\}$. Let $\hat{W}^{(l,0)}$ and $\hat{W}^{(l,1)}$ denote the weight matrices of $\hat{\mathcal{R}}^{(l)}$  and $\hat{\mathcal{S}}^{(l)}$ respectively. $\hat{W}_{i,j}^{(l,0)},\hat{W}_{i,j}^{(l,1)}\in [0,1]$. To make the whole Equation \ref{eq:discrete_logic} differentiable, we leverage the logical activation functions proposed by \cite{payani2019learning}:
\begin{equation}
\label{eq:conj_disj}
\textit{Conj}(\mathbf{h},
W_{i})=\prod_{j=1}^{n}F_{c}(\mathbf{h}_{j}, W_{i,j}), \;\;\;\;\;\;  \textit{Disj}(\mathbf{h}, W_{i})=1-\prod_{j=1}^{n}(1-F_{d}(\mathbf{h}_{j}, W_{i,j})),
\end{equation}
where $F_{c}(h,w)=1-w(1-h)$ and $F_{d}(h,w)=h\cdot w$. In Equation \ref{eq:conj_disj}, if  $\mathbf{h}$ and $W_{i}$ are both binary vectors, then $\textit{Conj}(\mathbf{h}, W_{i})=\bigwedge_{W_{i,j}=1}\mathbf{h}_{j}$ and $\textit{Disj}(\mathbf{h}, W_{i})=\bigvee_{W_{i,j}=1}\mathbf{h}_{j}$. $F_c(h,w)$ and $F_d(h,w)$ decide how much $\mathbf{h}_j$ would affect the operation according to $W_{i,j}$. If $W_{i,j}=0$, $\mathbf{h}_j$ would have no effect on the operation.
After using continuous weights and logical activation functions, the nodes in $\hat{\mathcal{R}}^{(l)}$ and $\hat{\mathcal{S}}^{(l)}$ are defined as follows:
\begin{equation}
\label{eq:conj_hat_disj_hat}
\hat{\mathbf{r}}_{i}^{(l)} =\textit{Conj}(\hat{\mathbf{u}}^{(l-1)}, \hat{W}_{i}^{(l,0)}), \;\;\;\;\;\; 
\hat{\mathbf{s}}_{i}^{(l)}=\textit{Disj}(\hat{\mathbf{u}}^{(l-1)}, \hat{W}_{i}^{(l,1)})
\end{equation}
Now the whole logical layer is differentiable and can be trained by gradient descent. However, the above logical activation functions suffer from the serious vanishing gradient problem. The main reason can be found by analyzing the partial derivative of each node w.r.t. its directly connected weights and w.r.t. its directly connected nodes as follows:
\begin{equation}
\label{eq:derivative_rw_ru}
\frac{\partial \hat{\mathbf{r}}_{i}^{(l)}}{\partial \hat{W}_{i,j}^{(l,0)}}
=(\hat{\mathbf{u}}_{j}^{(l-1)}-1) \cdot \prod_{k \neq j}F_c(\hat{\mathbf{u}}_{k}^{(l-1)}, \hat{W}_{i,k}^{(l,0)}),  \;\;\;\;\;
\frac{\partial \hat{\mathbf{r}}_{i}^{(l)}}{\partial \hat{\mathbf{u}}_{j}^{(l-1)}}
=\hat{W}_{i,j}^{(l,0)} \cdot \prod_{k \neq j}F_c(\hat{\mathbf{u}}_{k}^{(l-1)}, \hat{W}_{i,k}^{(l,0)})
\end{equation}
Since $\hat{\mathbf{u}}_{k}^{(l-1)}$ and $\hat{W}_{i,k}^{(l,0)}$ are in the range $[0, 1]$, the values of $F_c(\cdot)$ in Equation \ref{eq:derivative_rw_ru} are in the range $[0, 1]$ as well. If $\mathbf{n}_{l-1}$ is large and most of the values of $F_c(\cdot)$ are not 1, then the derivative is close to 0 due to the multiplications (See Appendix \ref{vanishing_gradient_problem} for the analysis of $\hat{\mathbf{s}}_{i}^{(l)}$). \cite{wang2020transparent} tries to use weight initialization to make $F_c(\cdot)$ close to 1 at the beginning. However, when dealing with hundreds of features, the vanishing gradient problem is still inevitable.

\noindent \textbf{Improved Logical Activation Functions.}
We found that using the multiplications to simulate the logical operations in Equation \ref{eq:conj_disj} is the main reason for vanishing gradients and propose an improved design of logical activation functions. One straightforward idea is to convert multiplications into additions using logarithm, e.g., $\log(\prod_{j=1}^{n}F_{c}(\mathbf{h}_{j}, W_{i,j}))=\sum_{j=1}^{n}\log(F_{c}(\mathbf{h}_{j}, W_{i,j}))$. However, after taking the logarithm, the logical activation functions in Equation \ref{eq:conj_disj} cannot keep the characteristics of logical operations any more, and the ranges of $Conj(\cdot)$ and $Disj(\cdot)$ are not $[0,1]$. To deal with this problem, we need a projection function to fix it. Apparently, the inverse function of $\log(x)$, i.e., $e^x$, is not suitable.

For the projection function $g$, three conditions must be satisfied: (\romannumeral1) $g(0)=e^0.$ (\romannumeral2) $\lim_{x\rightarrow -\infty}g(x)=\lim_{x\rightarrow -\infty }e^x=0.$ (\romannumeral3) $\lim_{x\rightarrow -\infty}\frac{e^x}{g(x)}=0.$ Condition (\romannumeral1) and (\romannumeral2) aim to keep the range and tendency of logical activation functions. Condition (\romannumeral3) aims to lower the speed of approaching zero when $x\rightarrow -\infty$.
In this work, we choose $g(x)=\frac{-1}{-1+x}$ as the projection function, and the improvement of logical activation functions can be summarized as the function $\mathbb{P}(v)=\frac{-1}{-1+\log(v)}$. The improved conjunction function $\textit{Conj}_{+}$ and disjunction function $\textit{Disj}_{+}$ are given by:
\begin{equation}
\label{eq:conj+_disj+}
\textit{Conj}_{+}(\mathbf{h},
W_{i})= \mathbb{P}(\prod_{j=1}^{n}(F_{c}(\mathbf{h}_{j}, W_{i,j})+\epsilon)), \;\; \textit{Disj}_{+}(\mathbf{h}, W_{i})=1-\mathbb{P}(\prod_{j=1}^{n}(1-F_{d}(\mathbf{h}_{j}, W_{i,j})+\epsilon))
\end{equation}
where $\epsilon$ is a small constant, e.g., $10^{-10}$. The improved logical activation functions can avoid the vanishing gradient problem in most scenarios and are much more scalable than the originals.
Moreover, considering that $\frac{d \mathbb{P}(v)}{d v}=\frac{\mathbb{P}^2(v)}{v}$, when $n$ in Equation \ref{eq:conj+_disj+} is extremely large, $\frac{d \mathbb{P}(v)}{d v}$ may be very close to 0 due to $\mathbb{P}^2(v)$. One trick to deal with it is replacing $\frac{\mathbb{P}^2(v)}{v}$ with $\frac{\mathbb{P}(\mathbb{P}^2(v))}{v}$ for $\mathbb{P}$ can lower the speed of approaching 0 while keeping the value range and tendency.
\subsection{Binarization Layer}
The binarization layer is mainly used to divide the continuous feature values into several bins. By combining one binarization layer and one logical layer, we can automatically choose the appropriate bins for feature discretization (binarization), i.e., binarizing features in an end-to-end way.

For the $j$-th continuous feature to be binarized, there are $k$ lower bounds ($\mathcal{T}_{j,1}, \dots, \mathcal{T}_{j,k}$) and $k$ upper bounds ($\mathcal{H}_{j,1}, \dots, \mathcal{H}_{j,k}$). All these bounds are randomly selected (e.g., from uniform distribution) in the value range of the $j$-th continuous feature, and these bounds are not trainable. When inputting one continuous feature vector $\vc$, the binarization layer will check if $\vc_{j}$ satisfies the bounds and get the following binary vector:
\begin{equation}
Q_{j} = [q(\vc_{j}-\mathcal{T}_{j,1}),\dots, q(\vc_{j}-\mathcal{T}_{j,k}),q(\mathcal{H}_{j,1}-\vc_{j}),\dots, q(\mathcal{H}_{j,k}-\vc_{j})],
\end{equation}
where $q(x)= \1_\mathrm{x>0}$. If we input the $i$-th instance, i.e., $\vc=C_i$, then $\bar{C}_i=Q_1\oplus Q_2 \dots \oplus Q_{m}$ and $\mathbf{u}^{(0)}=\bar{C}_i\oplus Bi$.
After inputting $\mathbf{u}^{(0)}$ to the logical layer $\mathcal{U}^{(1)}$, the edge connections between $\mathcal{U}^{(1)}$ and $\mathcal{U}^{(0)}$ indicate the choice of bounds (bins). For example, if $\mathbf{r}_{i}^{(1)}$ is connected to the nodes corresponding to $q(\vc_{j}-\mathcal{T}_{j,1})$ and $q(\mathcal{H}_{j,2}-\vc_{j})$, then $\mathbf{r}_{i}^{(1)}$ contains the bin $(\mathcal{T}_{j,1}<\vc_{j})\wedge(\vc_{j}<\mathcal{H}_{j,2})$. If we replace $\mathbf{r}_{i}^{(1)}$ with $\mathbf{s}_{i}^{(1)}$ in the example, we can get $(\mathcal{T}_{j,1}<\vc_{j})\vee(\vc_{j}<\mathcal{H}_{j,2})$. It should be noted that, in practice, if $\mathcal{T}_{j,1}\geq \mathcal{H}_{j,2}$, then $\mathbf{r}_{i}^{(1)}=0$, and if $\mathcal{T}_{j,1}< \mathcal{H}_{j,2}$, then $\mathbf{s}_{i}^{(1)}=1$. When using the continuous version, the weights of logical layers are trainable, which means we can choose bounds in an end-to-end way. For the number of bounds is $2k$ times of features, which could be large, only logical layers with improved logical activation functions are capable of choosing the bounds.
\subsection{Gradient Grafting}
\label{section:gradient_grafting}
Although RRL can be differentiable with the continuous logical layers, it is challenging to search for a discrete solution in a continuous space \citep{qin2020binary}. One commonly used method to tackle this problem is the Straight-Through Estimator (STE) \citep{courbariaux2016binarized}. The STE method needs gradients at discrete points to update the parameters. However, the gradients of RRL at discrete points have no useful information in most cases (See Appendix \ref{gradients_at_discrete_points}). Therefore STE is not suitable for RRL. Other methods like ProxQuant \citep{bai2018proxquant} and Random Binarization \citep{wang2020transparent} cannot directly optimize for the discrete model and be scalable at the same time.

Inspired by plant grafting, we propose a new training method, called Gradient Grafting, that can effectively train RRL. In stem grafting, one plant is selected for its roots, i.e., rootstock, and the other plant is selected for its stems, i.e., scion. By grafting, we obtain a new plant with the advantages of both two plants. In Gradient Grafting, the gradient of the loss function w.r.t. the output of discrete model is the scion, and the gradient of the output of continuous model w.r.t. the parameters of continuous model is the rootstock. Specifically, let $\theta$ denote the parameter vector and $\theta^t$ denote the parameter vector at step $t$. $q(\vx)= \1_{\vx>0.5}$ is the binarization function that binarizes each dimension of $\vx$ with 0.5 as the threshold. Let $\hat{Y}$ and $\bar{Y}$ denote the output of the continuous model $\hat{\mathcal{F}}$ and discrete model $\mathcal{F}$ respectively, then $\hat{Y}=\hat{\mathcal{F}}(\theta^{t},X)$, $\bar{Y}=\mathcal{F}(q(\theta^{t}),X)$. The parameters update with Gradient Grafting is formulated by:
\begin{equation}
\theta^{t+1}=\theta^{t}-\eta \frac{\partial \mathcal{L}(\bar{Y})}{\partial \bar{Y}}\cdot \frac{\partial \hat{Y}}{\partial \theta^{t}},
\end{equation}
where $\eta$ is the learning rate and $\mathcal{L}(\cdot)$ is the loss function. One simplified computation graph of Gradient Grafting is shown in Figure \ref{fig:gradient_grafting} for intuitive understanding.

Gradient Grafting can directly optimize the loss of discrete models and use the gradient information at both continuous and discrete points, which overcomes the problems occurring in RRL training when using other gradient-based discrete model training methods. The convergence of Gradient Grafting is verified in the experiments (See Figure \ref{fig:training_loss}).
\subsection{Model Interpretation}
After training with Gradient Grafting, the discrete RRL can be used for testing and interpretation. RRL is easy to interpret, for we can simply consider it as a feature learner and a linear classifier. The binarization layer and logical layers are the feature learner, and they use logical rules to build and describe the new features. The linear classifier, i.e., the linear layer, makes decisions based on the new features. We can first find the important new features by the weights of the linear layer, then understand each new feature by analyzing its corresponding rule. 
One advantage of RRL is that it can be easily adjusted by the practitioners to obtain a trade-off between the classification accuracy and model complexity. 
Therefore, RRL can satisfy the requirements from different tasks and scenarios. 
There are several ways to limit the model complexity of RRL. First, we can reduce the number of logical layers in RRL, i.e., the depth of RRL, and the number of nodes in each logical layer, i.e., the width of RRL. 
Second, the L1/L2 regularization can be used during training to search for an RRL with shorter rules. The coefficient of the regularization term in the loss function can be considered as a hyperparameter to restrict the model complexity. 
After training, the dead nodes detection and redundant rules elimination proposed by \cite{wang2020transparent} can also be used for better interpretability.

\begin{table*}[!t]
  \caption{5-fold cross validated F1 score (\%) of comparing models on all 13 datasets. $^*$ represents that RRL significantly outperforms all the compared interpretable models (t-test with p < 0.01). The first seven models are interpretable models, while the last five are complex models.}
  \label{tab:accuracy}
\resizebox{1.\linewidth}{!}{
  \begin{tabular}{c|ccccccc|ccccc}
    \toprule
    Dataset & \textbf{RRL} & C4.5 & CART & SBRL & CORELS & CRS & LR & SVM & PLNN(MLP) & RF & LightGBM & XGBoost\\
\midrule
adult & 80.72 & 77.77 & 77.06 & 79.88 & 70.56 & \textbf{80.95} & 78.43 & 63.63 & 73.55 & 79.22 & 80.36 & 80.64\\
bank-marketing & \textbf{76.32}$^{*}$ & 71.24 & 71.38 & 72.67 & 66.86 & 73.34 & 69.81 & 66.78 & 72.40 & 72.67 & 75.28 & 74.71\\
banknote & \textbf{100.00}$^{*}$ & 98.45 & 97.85 & 94.44 & 98.49 & 94.93 & 98.82 & \textbf{100.00} & \textbf{100.00} & 99.40 & 99.48 & 99.55\\
chess & 78.83 & 79.90 & 79.15 & 26.44 & 24.86 & 80.21 & 33.06 & 79.58 & 77.85 & 75.00 & 80.58 & \textbf{80.66}\\
connect-4 & \textbf{71.23}$^{*}$ & 61.66 & 61.24 & 48.54 & 51.72 & 65.88 & 49.87 & 69.85 & 64.55 & 62.72 & 70.53 & 70.65\\
letRecog & 96.15$^{*}$ & 88.20 & 87.62 & 64.32 & 61.13 & 84.96 & 72.05 & 95.57 & 92.34 & \textbf{96.59} & 96.51 & 96.38\\
magic04 & 86.33$^{*}$ & 82.44 & 81.20 & 82.52 & 77.37 & 80.87 & 75.72 & 79.43 & 83.07 & 86.48 & 86.67 & \textbf{86.69}\\
tic-tac-toe & \textbf{100.00} & 91.70 & 94.21 & 98.39 & 98.92 & 99.77 & 98.12 & 98.07 & 98.26 & 98.37 & 99.89 & 99.89\\
wine & 98.23 & 95.48 & 94.39 & 95.84 & 97.43 & 97.78 & 95.16 & 96.05 & 76.07 & 98.31 & \textbf{98.44} & 97.78\\
\midrule
activity & 98.17 & 94.24 & 93.35 & 11.34 & 51.61 & 5.05 & 98.47 & 98.67 & 98.27 & 97.80 & \textbf{99.41} & 99.38\\
dota2 & \textbf{60.12}$^{*}$ & 52.08 & 51.91 & 34.83 & 46.21 & 56.31 & 59.34 & 57.76 & 59.46 & 57.39 & 58.81 & 58.53\\
facebook & \textbf{90.27}$^{*}$ & 80.76 & 81.50 & 31.16 & 34.93 & 11.38 & 88.62 & 87.20 & 89.43 & 87.49 & 85.87 & 88.90\\
fashion & 89.01$^{*}$ & 80.49 & 79.61 & 47.38 & 38.06 & 66.92 & 84.53 & 84.46 & 89.36 & 88.35 & \textbf{89.91} & 89.82\\
\midrule
\textbf{AvgRank} & 2.77 & 8.23 & 8.92 & 9.31 & 9.92 & 7.08 & 7.92 & 6.77 & 5.77 & 5.38 & 2.77 & \textbf{2.69}\\
  \bottomrule
\end{tabular}
}
\end{table*}
\section{Experiments}
In this section, we conduct experiments to evaluate the proposed model and answer the following questions: (\romannumeral1) How is the classification performance of RRL? (\romannumeral2) How is the model complexity of RRL? (\romannumeral3) How is the convergence of Gradient Grafting compared to other gradient-based discrete model training methods? (\romannumeral4) How is the scalability of the improved logical activation functions? 
\subsection{Dataset Description and Experimental Settings}
\label{section:dataset_desc_and_exp_settings}
\textbf{Dataset Description.} We took nine small and four large public datasets to conduct our experiments, all of which are often used to test classification performance and model interpretability \citep{Dua:2019,xiao2017:online,anguita2013public,rozemberczki2019multiscale}. Appendix \ref{appendix:data_propertity} summarizes the statistics of these 13 datasets.
Together they show the data diversity, ranging from 178 to 102944 instances, from 2 to 26 classes, and from 4 to 4714 original features. See Appendix \ref{appendix:code_and_data} for licenses.

\noindent \textbf{Performance Measurement.}
We adopt the F1 score (Macro) as the classification performance metric since some of the data sets are imbalanced, i.e., the numbers of different classes are quite different. We adopt 5-fold cross-validation to evaluate the classification performance more fairly.
The average rank of each model is also adopted for comparisons of classification performance over all the data sets \citep{demvsar2006statistical}.
Considering that reused structures exist in rule-based models, e.g., one branch in Decision Tree can correspond to several rules, we use the total number of edges instead of the total length of all rules as the metric of model complexity for rule-based models. See Appendix \ref{appendix:experimental_setting} for details about the \textbf{experiment environment} and \textbf{parameter settings} of all models.

\subsection{Classification Performance}
\label{section:classification_performance}
We compare the classification F1 score (Macro) of RRL with six interpretable models and five complex models. C4.5 \citep{Quinlan:1993:CPM:583200}, CART \citep{breiman2017classification}, Scalable Bayesian Rule Lists (SBRL) \citep{yang2017scalable}, Certifiably Optimal Rule Lists (CORELS) \citep{angelino2017learning}, and Concept Rule Sets (CRS) \citep{wang2020transparent} are rule-based models. Logistic Regression (LR) \citep{kleinbaum2002logistic} is a linear model. These six models are considered interpretable models. Piecewise Linear Neural Network (PLNN) \citep{chu2018exact}, Support Vector Machines (SVM) \citep{scholkopf2001learning}, Random Forest \citep{breiman2001random}, LightGBM \citep{ke2017lightgbm}, and XGBoost \citep{chen2016xgboost} are considered complex models. PLNN is a Multilayer Perceptron (MLP) that adopts piecewise linear activation functions, e.g., ReLU \citep{nair2010rectified}. RF, LightGBM, and XGBoost are ensemble models. See Appendix \ref{appendix:experimental_setting} for the parameters tuning.

The results are shown in Table \ref{tab:accuracy}, and the first nine data sets are small data sets while the last four are large data sets. We can observe that RRL performs well on almost all the data sets and gets the best results on 6 data sets. The two-tailed Student’s t-test (p<0.01) is used for significance testing, and we can observe that RRL significantly outperforms all the compared interpretable models on 8 out of 13 data sets. The average rank of RRL is also the top three among all the models. Only two complex models that use hundreds of estimators, i.e., XGBoost and LightGBM, have comparable results with RRL. Comparing RRL with LR and other rule-based models, we can see RRL can fit both linear and non-linear data well. CRS performs well on small data sets but fails on large datasets due to the limitation of its logical activation functions and training method. Good results on both small and large data sets verify RRL has good scalability.
Moreover, SBRL and CRS do not perform well on continuous feature data sets like \textit{letRecog} and \textit{magic04} for they need preprocessing to discretize continuous features, which may bring bias to the data sets. On the contrary, RRL overcomes this problem by discretizing features end-to-end.
\begin{figure*}
    \centering
    \includegraphics[width=0.95\textwidth]{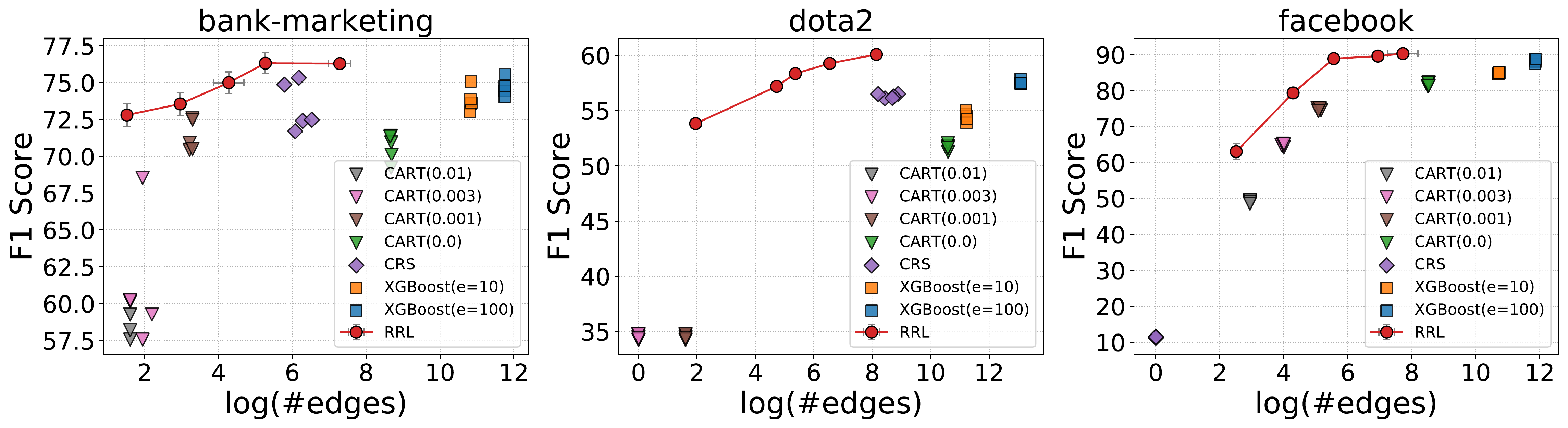}
    \vspace{-4pt}
    \caption{Scatter plot of F1 score against log(\#edges) for RRL and baselines on three datasets (see Appendix \ref{appendix:model_complexity} for other datasets).}
    \label{fig:model_complexity}
\end{figure*}
\subsection{Model Complexity}
Interpretable models seek to keep low model complexity while ensuring high accuracy. 
To show the relationships between accuracy and model complexity of RRL and baselines, we draw scatter plots of F1 score against log(\#edges) for rule-based models in Figure \ref{fig:model_complexity} (see Appendix \ref{appendix:model_complexity} for other data sets). 
The baselines are typical models in different categories of methods with good trade-offs between accuracy and model complexity.
For RRL, the legend markers and error bars indicate means and standard deviations, respectively, of F1 score and log(\#edges) across cross-validation folds.
For baseline models, each point represents an evaluation of one model, on one fold, with one parameter setting.
Therefore, in Figure \ref{fig:model_complexity}, the closer its corresponding points are to the upper left corner, the better one model is.
To obtain RRL with different model complexities, we tune the depth and width of RRL and the coefficient of L2 regularization term. 
The value in CART($\cdot$), e.g., CART(0.03), denotes the complexity parameter used for Minimal Cost-Complexity Pruning \citep{breiman2017classification}, and a higher value corresponds to a simpler tree. We also show results of XGBoost with 10 and 100 estimators.

In Figure \ref{fig:model_complexity}, on both small and large data sets, we can observe that if we connect the results of RRL, it will become a boundary that separating the upper left corner from other models. In other words, if RRL has a close model complexity with one baseline, then the F1 score of RRL will be higher. If the F1 score of RRL is close to one baseline, then the model complexity of RRL will be lower. It indicates that RRL can make better use of rules than rule-based models using heuristic and ensemble methods in most cases.
The results in Figure \ref{fig:model_complexity} also verify that we can adjust the model complexity of RRL by setting the model structure and the coefficient of L2 regularization term. 
In this way, the practitioners are able to select an RRL with suitable classification performance and model complexity for different scenarios, which is crucial for practical applications of interpretable models.
\begin{figure*}
    \centering
    \includegraphics[width=0.95\textwidth]{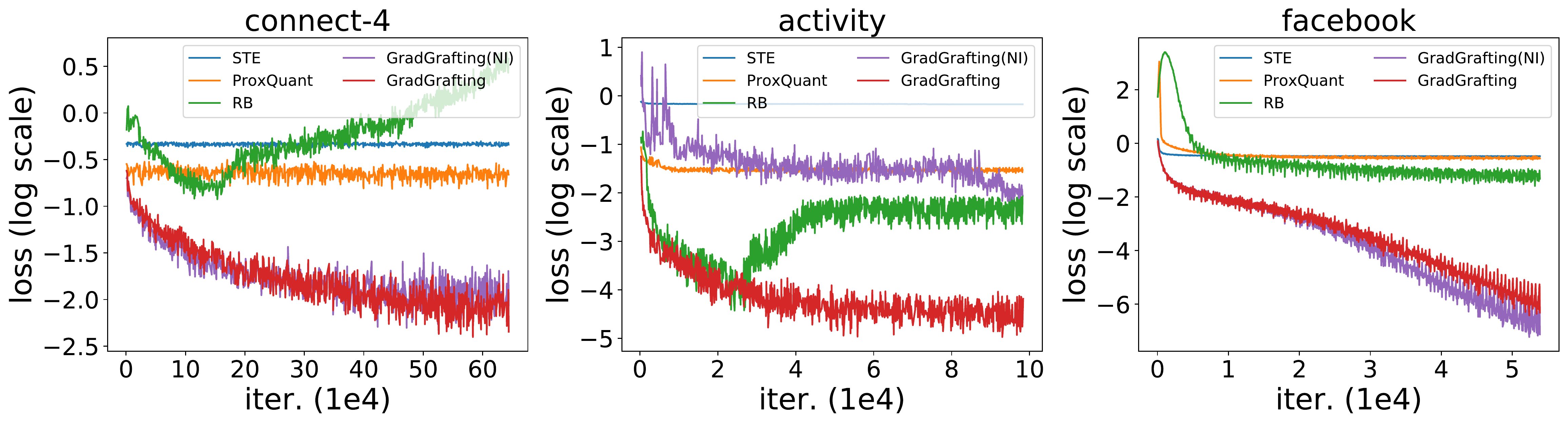}
    \vspace{-5pt}
    \caption{Training loss of three compared discrete model training methods and Gradient Grafting with or without improved logical activation functions on three data sets.}
    \label{fig:training_loss}
\end{figure*}
\begin{figure*}
\centering
  \begin{subfigure}[c]{0.3\linewidth}
    \includegraphics[width=\linewidth]{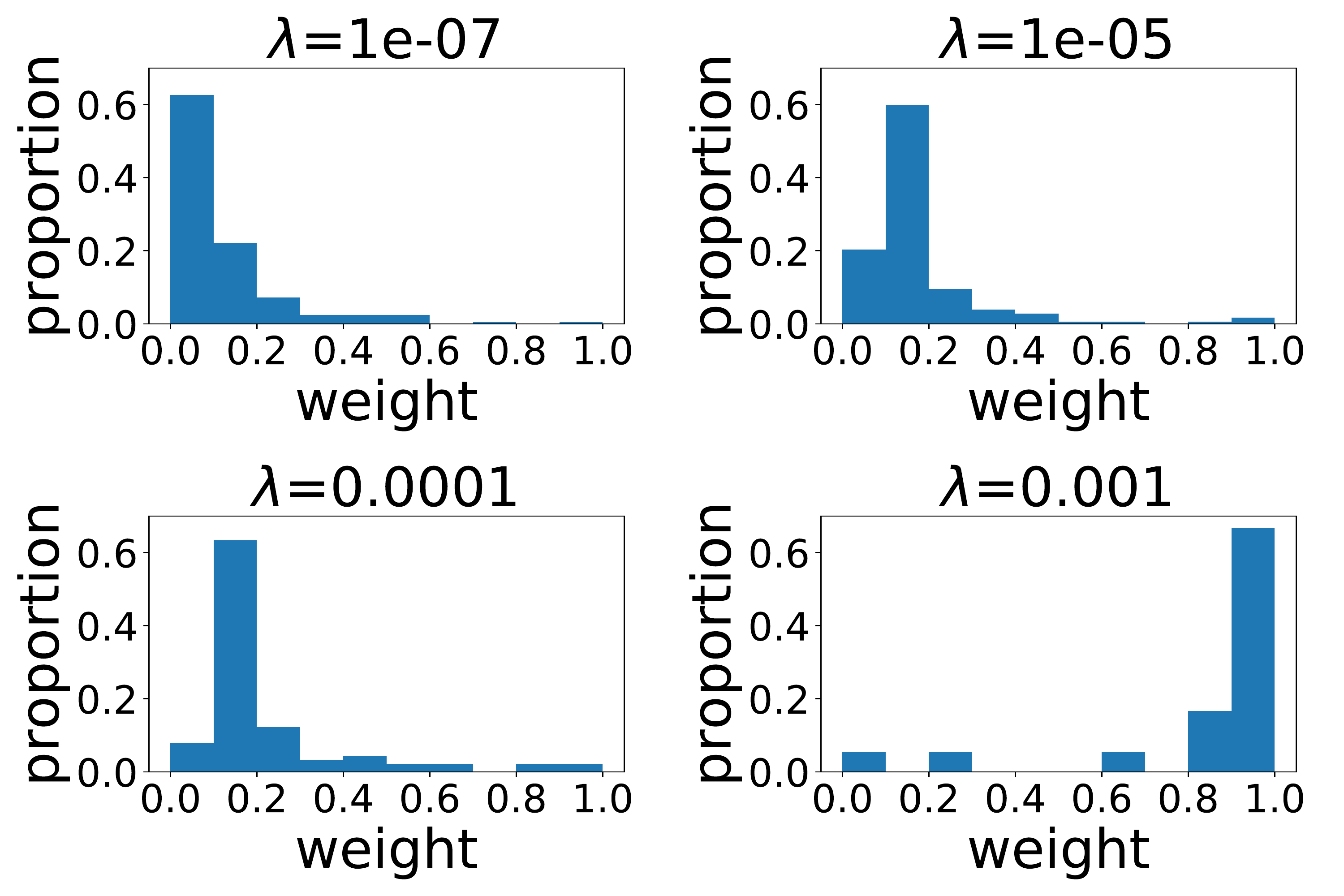}
    \vspace{-15pt}
    \caption{}
    \label{fig:weights_distribution}
  \end{subfigure}
  \hfill
  \begin{subfigure}[c]{0.65\linewidth}
    \includegraphics[width=\linewidth]{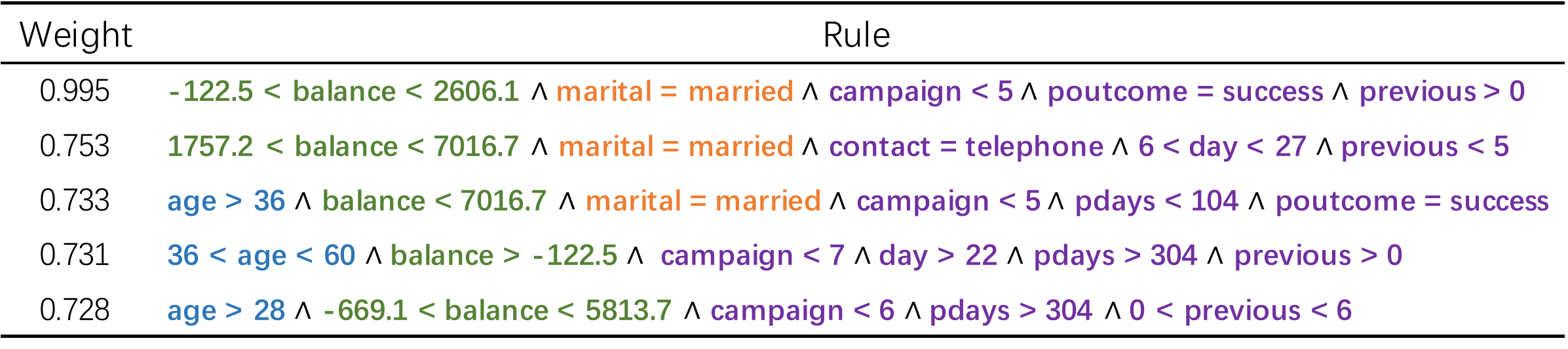}
    \caption{}
    \label{fig:bank-marketing}
  \end{subfigure}
\vspace{-1pt}
\caption{(a) The distribution of weights in the linear layer of RRLs trained on the \textit{bank-marketing} data set with the same model structure but different $\lambda$, where $\lambda$ is the coefficient of the L2 regularization term. (b) Logical rules obtained from RRL trained on the \textit{bank-marketing} data set.}
\label{fig:selector}
\end{figure*}

\subsection{Ablation Study}
\textbf{Training Method for Discrete Model.} To show the effectiveness of Gradient Grafting for training RRL, we compare it with other representative gradient-based discrete model training methods, i.e., STE \citep{cour2015bconnect,courbariaux2016binarized}, ProxQuant \citep{bai2018proxquant} and RB \citep{wang2020transparent}, by training RRL with the same structure. Hyperparameters are set to be the same for each method except exclusive hyperparameters, e.g., random binarization rate for RB, are fine-tuned. The training loss of the compared discrete model training methods and Gradient Grafting are shown in Figure \ref{fig:training_loss}, and we can see that the convergence of Gradient Grafting is faster and stabler than other methods on all data sets. As we mentioned in Section \ref{section:gradient_grafting}, RRL has little useful gradient information at discrete points, thus RRL trained by STE cannot converge. Due to the difference between discrete and continuous RRL, RRL trained by ProxQuant and RB cannot converge well as well.

\noindent \textbf{Improved Logical Activation Functions.} We also compare RRL trained by Gradient Grafting with or without improved logical activation functions. The results are also shown in Figure \ref{fig:training_loss}, and GradGrafting(NI) represents RRL using original logical activation functions instead of improved logical activation functions. We can observe that the original activation functions work well on small data sets but fail on the large data set \textit{activity} while the improved activation functions work well on all data sets, which means the improved logical activation functions make RRL more scalable. It should be noted that GradGrafting(NI) works well on the large data set \textit{facebook}. The reason is \textit{facebook} is a very sparse data set, and the number of 1 in each binary feature vector is less than 30 (See Appendix \ref{vanishing_gradient_problem} for detailed analyses).
\subsection{Case Study}
We show how the learned RRL looks like by case studies. Take the \textit{bank-marketing} data set as an example. We first show the distribution of weights in the linear layer of the trained RRLs in Figure \ref{fig:weights_distribution}. 
Each weight in the linear layer corresponds to one rule. 
For a better viewing experience, we show the normalized absolute values of weights.
The model structures of these RRLs are the same, but different coefficients of the L2 regularization term, denoted by $\lambda$, lead to different model complexities.
We can observe that, when $\lambda$ is small, which means the RRL is more complex, there are many rules with small weights. These small weight rules are mainly used to slightly adjust the outputs. Hence, they make the RRL more accurate but less interpretable. However, in practice, we can ignore these small weight rules and only focus on rules with large weights first. After analyzing rules with large weights and having a better understanding of the learned RRL and the data set, we can then understand those less important rules gradually.
When $\lambda$ is large, the number of rules is small, and we can directly understand the whole RRL rather than understanding RRL step by step.

In Figure \ref{fig:bank-marketing}, we show the learned rules, with high weights, of one RRL trained on the \textit{bank-marketing} data set (see Appendix \ref{appendix:case_study} for the \textit{fashion} data set). These rules are used to predict if the client will subscribe a term deposit by telesales. Different types of features are marked in different colors, e.g., purple for previous behaviours of the bank. We can clearly see that middle-aged married persons with low balance are more likely to subscribe a deposit, and the previous behaviour of the bank would also affect the client. Then the bank can change its strategies according to these rules.

\section{Conclusion and Future Work}
\label{section:conclusion}
We propose a new scalable classifier, named Rule-based Representation Learner (RRL), that can automatically learn interpretable rules for data representation and classification. For the particularity of RRL, we propose a new gradient-based discrete model training method, i.e., Gradient Grafting, that directly optimizes the discrete model. We also propose an improved design of logical activation functions to increase the scalability of RRL and make RRL capable of discretizing the continuous features end-to-end. Our experimental results show that RRL enjoys both high classification performance and low model complexity on data sets with different scales. For the current design of RRL is limited to structured data, we will extend RRL to suit more unstructured data as future work.

\bibliographystyle{ACM-Reference-Format}
\bibliography{sample-base}


\begin{thebibliography}{40}


\ifx \showCODEN    \undefined \def \showCODEN     #1{\unskip}     \fi
\ifx \showDOI      \undefined \def \showDOI       #1{#1}\fi
\ifx \showISBNx    \undefined \def \showISBNx     #1{\unskip}     \fi
\ifx \showISBNxiii \undefined \def \showISBNxiii  #1{\unskip}     \fi
\ifx \showISSN     \undefined \def \showISSN      #1{\unskip}     \fi
\ifx \showLCCN     \undefined \def \showLCCN      #1{\unskip}     \fi
\ifx \shownote     \undefined \def \shownote      #1{#1}          \fi
\ifx \showarticletitle \undefined \def \showarticletitle #1{#1}   \fi
\ifx \showURL      \undefined \def \showURL       {\relax}        \fi
\providecommand\bibfield[2]{#2}
\providecommand\bibinfo[2]{#2}
\providecommand\natexlab[1]{#1}
\providecommand\showeprint[2][]{arXiv:#2}

\bibitem[\protect\citeauthoryear{Angelino, Larus-Stone, Alabi, Seltzer, and
  Rudin}{Angelino et~al\mbox{.}}{2017}]%
        {angelino2017learning}
\bibfield{author}{\bibinfo{person}{Elaine Angelino}, \bibinfo{person}{Nicholas
  Larus-Stone}, \bibinfo{person}{Daniel Alabi}, \bibinfo{person}{Margo
  Seltzer}, {and} \bibinfo{person}{Cynthia Rudin}.}
  \bibinfo{year}{2017}\natexlab{}.
\newblock \showarticletitle{Learning certifiably optimal rule lists for
  categorical data}.
\newblock \bibinfo{journal}{\emph{The Journal of Machine Learning Research}}
  \bibinfo{volume}{18}, \bibinfo{number}{1} (\bibinfo{year}{2017}),
  \bibinfo{pages}{8753--8830}.
\newblock


\bibitem[\protect\citeauthoryear{Anguita, Ghio, Oneto, Parra, and
  Reyes-Ortiz}{Anguita et~al\mbox{.}}{2013}]%
        {anguita2013public}
\bibfield{author}{\bibinfo{person}{Davide Anguita}, \bibinfo{person}{Alessandro
  Ghio}, \bibinfo{person}{Luca Oneto}, \bibinfo{person}{Xavier Parra}, {and}
  \bibinfo{person}{Jorge~Luis Reyes-Ortiz}.} \bibinfo{year}{2013}\natexlab{}.
\newblock \showarticletitle{A public domain dataset for human activity
  recognition using smartphones.}. In \bibinfo{booktitle}{\emph{Esann}}.
\newblock


\bibitem[\protect\citeauthoryear{Bai, Wang, and Liberty}{Bai
  et~al\mbox{.}}{2018}]%
        {bai2018proxquant}
\bibfield{author}{\bibinfo{person}{Yu Bai}, \bibinfo{person}{Yu-Xiang Wang},
  {and} \bibinfo{person}{Edo Liberty}.} \bibinfo{year}{2018}\natexlab{}.
\newblock \showarticletitle{Proxquant: Quantized neural networks via proximal
  operators}.
\newblock \bibinfo{journal}{\emph{arXiv preprint arXiv:1810.00861}}
  (\bibinfo{year}{2018}).
\newblock


\bibitem[\protect\citeauthoryear{Breiman}{Breiman}{2001}]%
        {breiman2001random}
\bibfield{author}{\bibinfo{person}{Leo Breiman}.}
  \bibinfo{year}{2001}\natexlab{}.
\newblock \showarticletitle{Random forests}.
\newblock \bibinfo{journal}{\emph{Machine learning}} \bibinfo{volume}{45},
  \bibinfo{number}{1} (\bibinfo{year}{2001}), \bibinfo{pages}{5--32}.
\newblock


\bibitem[\protect\citeauthoryear{Breiman}{Breiman}{2017}]%
        {breiman2017classification}
\bibfield{author}{\bibinfo{person}{Leo Breiman}.}
  \bibinfo{year}{2017}\natexlab{}.
\newblock \bibinfo{booktitle}{\emph{Classification and regression trees}}.
\newblock \bibinfo{publisher}{Routledge}.
\newblock


\bibitem[\protect\citeauthoryear{Chen and Guestrin}{Chen and Guestrin}{2016}]%
        {chen2016xgboost}
\bibfield{author}{\bibinfo{person}{Tianqi Chen} {and} \bibinfo{person}{Carlos
  Guestrin}.} \bibinfo{year}{2016}\natexlab{}.
\newblock \showarticletitle{Xgboost: A scalable tree boosting system}. In
  \bibinfo{booktitle}{\emph{Proceedings of the 22nd acm sigkdd international
  conference on knowledge discovery and data mining}}.
  \bibinfo{pages}{785--794}.
\newblock


\bibitem[\protect\citeauthoryear{Chu, Hu, Hu, Wang, and Pei}{Chu
  et~al\mbox{.}}{2018}]%
        {chu2018exact}
\bibfield{author}{\bibinfo{person}{Lingyang Chu}, \bibinfo{person}{Xia Hu},
  \bibinfo{person}{Juhua Hu}, \bibinfo{person}{Lanjun Wang}, {and}
  \bibinfo{person}{Jian Pei}.} \bibinfo{year}{2018}\natexlab{}.
\newblock \showarticletitle{Exact and consistent interpretation for piecewise
  linear neural networks: A closed form solution}. In
  \bibinfo{booktitle}{\emph{SIGKDD}}. ACM, \bibinfo{pages}{1244--1253}.
\newblock


\bibitem[\protect\citeauthoryear{Cohen}{Cohen}{1995}]%
        {cohen1995fast}
\bibfield{author}{\bibinfo{person}{William~W Cohen}.}
  \bibinfo{year}{1995}\natexlab{}.
\newblock \showarticletitle{Fast effective rule induction}.
\newblock In \bibinfo{booktitle}{\emph{MLP}}. \bibinfo{publisher}{Elsevier},
  \bibinfo{pages}{115--123}.
\newblock


\bibitem[\protect\citeauthoryear{Courbariaux, Bengio, and David}{Courbariaux
  et~al\mbox{.}}{2015}]%
        {cour2015bconnect}
\bibfield{author}{\bibinfo{person}{Matthieu Courbariaux},
  \bibinfo{person}{Yoshua Bengio}, {and} \bibinfo{person}{Jean-Pierre David}.}
  \bibinfo{year}{2015}\natexlab{}.
\newblock \showarticletitle{Binaryconnect: Training deep neural networks with
  binary weights during propagations}. In \bibinfo{booktitle}{\emph{NeurIPS}}.
  \bibinfo{pages}{3123--3131}.
\newblock


\bibitem[\protect\citeauthoryear{Courbariaux, Hubara, Soudry, El-Yaniv, and
  Bengio}{Courbariaux et~al\mbox{.}}{2016}]%
        {courbariaux2016binarized}
\bibfield{author}{\bibinfo{person}{Matthieu Courbariaux}, \bibinfo{person}{Itay
  Hubara}, \bibinfo{person}{Daniel Soudry}, \bibinfo{person}{Ran El-Yaniv},
  {and} \bibinfo{person}{Yoshua Bengio}.} \bibinfo{year}{2016}\natexlab{}.
\newblock \showarticletitle{Binarized neural networks: Training deep neural
  networks with weights and activations constrained to+ 1 or-1}.
\newblock \bibinfo{journal}{\emph{arXiv preprint arXiv:1602.02830}}
  (\bibinfo{year}{2016}).
\newblock


\bibitem[\protect\citeauthoryear{Dem{\v{s}}ar}{Dem{\v{s}}ar}{2006}]%
        {demvsar2006statistical}
\bibfield{author}{\bibinfo{person}{Janez Dem{\v{s}}ar}.}
  \bibinfo{year}{2006}\natexlab{}.
\newblock \showarticletitle{Statistical comparisons of classifiers over
  multiple data sets}.
\newblock \bibinfo{journal}{\emph{Journal of Machine learning research}}
  \bibinfo{volume}{7}, \bibinfo{number}{Jan} (\bibinfo{year}{2006}),
  \bibinfo{pages}{1--30}.
\newblock


\bibitem[\protect\citeauthoryear{Doshi-Velez and Kim}{Doshi-Velez and
  Kim}{2017}]%
        {doshi2017towards}
\bibfield{author}{\bibinfo{person}{Finale Doshi-Velez} {and}
  \bibinfo{person}{Been Kim}.} \bibinfo{year}{2017}\natexlab{}.
\newblock \showarticletitle{Towards a rigorous science of interpretable machine
  learning}.
\newblock \bibinfo{journal}{\emph{arXiv preprint arXiv:1702.08608}}
  (\bibinfo{year}{2017}).
\newblock


\bibitem[\protect\citeauthoryear{Dua and Graff}{Dua and Graff}{2017}]%
        {Dua:2019}
\bibfield{author}{\bibinfo{person}{Dheeru Dua} {and} \bibinfo{person}{Casey
  Graff}.} \bibinfo{year}{2017}\natexlab{}.
\newblock \bibinfo{title}{{UCI} Machine Learning Repository}.
\newblock
\newblock
\urldef\tempurl%
\url{http://archive.ics.uci.edu/ml}
\showURL{%
\tempurl}


\bibitem[\protect\citeauthoryear{Frosst and Hinton}{Frosst and Hinton}{2017}]%
        {frosst2017distilling}
\bibfield{author}{\bibinfo{person}{Nicholas Frosst} {and}
  \bibinfo{person}{Geoffrey Hinton}.} \bibinfo{year}{2017}\natexlab{}.
\newblock \showarticletitle{Distilling a neural network into a soft decision
  tree}.
\newblock \bibinfo{journal}{\emph{arXiv preprint arXiv:1711.09784}}
  (\bibinfo{year}{2017}).
\newblock


\bibitem[\protect\citeauthoryear{Goodfellow, Bengio, Courville, and
  Bengio}{Goodfellow et~al\mbox{.}}{2016}]%
        {goodfellow2016deep}
\bibfield{author}{\bibinfo{person}{Ian Goodfellow}, \bibinfo{person}{Yoshua
  Bengio}, \bibinfo{person}{Aaron Courville}, {and} \bibinfo{person}{Yoshua
  Bengio}.} \bibinfo{year}{2016}\natexlab{}.
\newblock \bibinfo{booktitle}{\emph{Deep learning}}. Vol.~\bibinfo{volume}{1}.
\newblock \bibinfo{publisher}{MIT Press}.
\newblock


\bibitem[\protect\citeauthoryear{Hara and Hayashi}{Hara and Hayashi}{2016}]%
        {hara2016making}
\bibfield{author}{\bibinfo{person}{Satoshi Hara} {and} \bibinfo{person}{Kohei
  Hayashi}.} \bibinfo{year}{2016}\natexlab{}.
\newblock \showarticletitle{Making tree ensembles interpretable}.
\newblock \bibinfo{journal}{\emph{arXiv preprint arXiv:1606.05390}}
  (\bibinfo{year}{2016}).
\newblock


\bibitem[\protect\citeauthoryear{Irsoy, Y{\i}ld{\i}z, and Alpayd{\i}n}{Irsoy
  et~al\mbox{.}}{2012}]%
        {irsoy2012soft}
\bibfield{author}{\bibinfo{person}{Ozan Irsoy}, \bibinfo{person}{Olcay~Taner
  Y{\i}ld{\i}z}, {and} \bibinfo{person}{Ethem Alpayd{\i}n}.}
  \bibinfo{year}{2012}\natexlab{}.
\newblock \showarticletitle{Soft decision trees}. In
  \bibinfo{booktitle}{\emph{Proceedings of the 21st International Conference on
  Pattern Recognition (ICPR2012)}}. IEEE, \bibinfo{pages}{1819--1822}.
\newblock


\bibitem[\protect\citeauthoryear{Ishibuchi and Yamamoto}{Ishibuchi and
  Yamamoto}{2005}]%
        {ishibuchi2005rule}
\bibfield{author}{\bibinfo{person}{Hisao Ishibuchi} {and}
  \bibinfo{person}{Takashi Yamamoto}.} \bibinfo{year}{2005}\natexlab{}.
\newblock \showarticletitle{Rule weight specification in fuzzy rule-based
  classification systems}.
\newblock \bibinfo{journal}{\emph{IEEE transactions on fuzzy systems}}
  \bibinfo{volume}{13}, \bibinfo{number}{4} (\bibinfo{year}{2005}),
  \bibinfo{pages}{428--435}.
\newblock


\bibitem[\protect\citeauthoryear{Jang, Gu, and Poole}{Jang
  et~al\mbox{.}}{2016}]%
        {jang2016categorical}
\bibfield{author}{\bibinfo{person}{Eric Jang}, \bibinfo{person}{Shixiang Gu},
  {and} \bibinfo{person}{Ben Poole}.} \bibinfo{year}{2016}\natexlab{}.
\newblock \showarticletitle{Categorical reparameterization with
  gumbel-softmax}.
\newblock \bibinfo{journal}{\emph{arXiv preprint arXiv:1611.01144}}
  (\bibinfo{year}{2016}).
\newblock


\bibitem[\protect\citeauthoryear{Ke, Meng, Finley, Wang, Chen, Ma, Ye, and
  Liu}{Ke et~al\mbox{.}}{2017}]%
        {ke2017lightgbm}
\bibfield{author}{\bibinfo{person}{Guolin Ke}, \bibinfo{person}{Qi Meng},
  \bibinfo{person}{Thomas Finley}, \bibinfo{person}{Taifeng Wang},
  \bibinfo{person}{Wei Chen}, \bibinfo{person}{Weidong Ma},
  \bibinfo{person}{Qiwei Ye}, {and} \bibinfo{person}{Tie-Yan Liu}.}
  \bibinfo{year}{2017}\natexlab{}.
\newblock \showarticletitle{Lightgbm: A highly efficient gradient boosting
  decision tree}. In \bibinfo{booktitle}{\emph{Advances in Neural Information
  Processing Systems}}. \bibinfo{pages}{3146--3154}.
\newblock


\bibitem[\protect\citeauthoryear{Kingma and Ba}{Kingma and Ba}{2014}]%
        {kingma2014adam}
\bibfield{author}{\bibinfo{person}{Diederik~P Kingma} {and}
  \bibinfo{person}{Jimmy Ba}.} \bibinfo{year}{2014}\natexlab{}.
\newblock \showarticletitle{Adam: A method for stochastic optimization}.
\newblock \bibinfo{journal}{\emph{arXiv preprint arXiv:1412.6980}}
  (\bibinfo{year}{2014}).
\newblock


\bibitem[\protect\citeauthoryear{Kleinbaum, Dietz, Gail, Klein, and
  Klein}{Kleinbaum et~al\mbox{.}}{2002}]%
        {kleinbaum2002logistic}
\bibfield{author}{\bibinfo{person}{David~G Kleinbaum}, \bibinfo{person}{K
  Dietz}, \bibinfo{person}{M Gail}, \bibinfo{person}{Mitchel Klein}, {and}
  \bibinfo{person}{Mitchell Klein}.} \bibinfo{year}{2002}\natexlab{}.
\newblock \bibinfo{booktitle}{\emph{Logistic regression}}.
\newblock \bibinfo{publisher}{Springer}.
\newblock


\bibitem[\protect\citeauthoryear{Lakkaraju, Bach, and Leskovec}{Lakkaraju
  et~al\mbox{.}}{2016}]%
        {lakkaraju2016interpretable}
\bibfield{author}{\bibinfo{person}{Himabindu Lakkaraju},
  \bibinfo{person}{Stephen~H Bach}, {and} \bibinfo{person}{Jure Leskovec}.}
  \bibinfo{year}{2016}\natexlab{}.
\newblock \showarticletitle{Interpretable decision sets: A joint framework for
  description and prediction}. In \bibinfo{booktitle}{\emph{SIGKDD}}. ACM,
  \bibinfo{pages}{1675--1684}.
\newblock


\bibitem[\protect\citeauthoryear{Letham, Rudin, McCormick, Madigan,
  et~al\mbox{.}}{Letham et~al\mbox{.}}{2015}]%
        {letham2015interpretable}
\bibfield{author}{\bibinfo{person}{Benjamin Letham}, \bibinfo{person}{Cynthia
  Rudin}, \bibinfo{person}{Tyler~H McCormick}, \bibinfo{person}{David Madigan},
  {et~al\mbox{.}}} \bibinfo{year}{2015}\natexlab{}.
\newblock \showarticletitle{Interpretable classifiers using rules and bayesian
  analysis: Building a better stroke prediction model}.
\newblock \bibinfo{journal}{\emph{The Annals of Applied Statistics}}
  \bibinfo{volume}{9}, \bibinfo{number}{3} (\bibinfo{year}{2015}),
  \bibinfo{pages}{1350--1371}.
\newblock


\bibitem[\protect\citeauthoryear{Lipton}{Lipton}{2016}]%
        {lipton2016mythos}
\bibfield{author}{\bibinfo{person}{Zachary~C Lipton}.}
  \bibinfo{year}{2016}\natexlab{}.
\newblock \showarticletitle{The mythos of model interpretability}.
\newblock \bibinfo{journal}{\emph{arXiv preprint arXiv:1606.03490}}
  (\bibinfo{year}{2016}).
\newblock


\bibitem[\protect\citeauthoryear{Molnar}{Molnar}{2019}]%
        {molnar2019}
\bibfield{author}{\bibinfo{person}{Christoph Molnar}.}
  \bibinfo{year}{2019}\natexlab{}.
\newblock \bibinfo{booktitle}{\emph{Interpretable Machine Learning}}.
\newblock
\newblock
\shownote{\url{https://christophm.github.io/interpretable-ml-book/}.}


\bibitem[\protect\citeauthoryear{Murdoch, Singh, Kumbier, Abbasi-Asl, and
  Yu}{Murdoch et~al\mbox{.}}{2019}]%
        {murdoch2019interpretable}
\bibfield{author}{\bibinfo{person}{W~James Murdoch}, \bibinfo{person}{Chandan
  Singh}, \bibinfo{person}{Karl Kumbier}, \bibinfo{person}{Reza Abbasi-Asl},
  {and} \bibinfo{person}{Bin Yu}.} \bibinfo{year}{2019}\natexlab{}.
\newblock \showarticletitle{Interpretable machine learning: definitions,
  methods, and applications}.
\newblock \bibinfo{journal}{\emph{arXiv preprint arXiv:1901.04592}}
  (\bibinfo{year}{2019}).
\newblock


\bibitem[\protect\citeauthoryear{Nair and Hinton}{Nair and Hinton}{2010}]%
        {nair2010rectified}
\bibfield{author}{\bibinfo{person}{Vinod Nair} {and}
  \bibinfo{person}{Geoffrey~E Hinton}.} \bibinfo{year}{2010}\natexlab{}.
\newblock \showarticletitle{Rectified linear units improve restricted boltzmann
  machines}. In \bibinfo{booktitle}{\emph{ICML}}. \bibinfo{pages}{807--814}.
\newblock


\bibitem[\protect\citeauthoryear{Paszke, Gross, Massa, Lerer, Bradbury, Chanan,
  Killeen, Lin, Gimelshein, Antiga, et~al\mbox{.}}{Paszke
  et~al\mbox{.}}{2019}]%
        {paszke2019pytorch}
\bibfield{author}{\bibinfo{person}{Adam Paszke}, \bibinfo{person}{Sam Gross},
  \bibinfo{person}{Francisco Massa}, \bibinfo{person}{Adam Lerer},
  \bibinfo{person}{James Bradbury}, \bibinfo{person}{Gregory Chanan},
  \bibinfo{person}{Trevor Killeen}, \bibinfo{person}{Zeming Lin},
  \bibinfo{person}{Natalia Gimelshein}, \bibinfo{person}{Luca Antiga},
  {et~al\mbox{.}}} \bibinfo{year}{2019}\natexlab{}.
\newblock \showarticletitle{Pytorch: An imperative style, high-performance deep
  learning library}. In \bibinfo{booktitle}{\emph{Advances in neural
  information processing systems}}. \bibinfo{pages}{8026--8037}.
\newblock


\bibitem[\protect\citeauthoryear{Payani and Fekri}{Payani and Fekri}{2019}]%
        {payani2019learning}
\bibfield{author}{\bibinfo{person}{Ali Payani} {and} \bibinfo{person}{Faramarz
  Fekri}.} \bibinfo{year}{2019}\natexlab{}.
\newblock \showarticletitle{Learning algorithms via neural logic networks}.
\newblock \bibinfo{journal}{\emph{arXiv preprint arXiv:1904.01554}}
  (\bibinfo{year}{2019}).
\newblock


\bibitem[\protect\citeauthoryear{Qin, Gong, Liu, Bai, Song, and Sebe}{Qin
  et~al\mbox{.}}{2020}]%
        {qin2020binary}
\bibfield{author}{\bibinfo{person}{Haotong Qin}, \bibinfo{person}{Ruihao Gong},
  \bibinfo{person}{Xianglong Liu}, \bibinfo{person}{Xiao Bai},
  \bibinfo{person}{Jingkuan Song}, {and} \bibinfo{person}{Nicu Sebe}.}
  \bibinfo{year}{2020}\natexlab{}.
\newblock \showarticletitle{Binary neural networks: A survey}.
\newblock \bibinfo{journal}{\emph{Pattern Recognition}} (\bibinfo{year}{2020}),
  \bibinfo{pages}{107281}.
\newblock


\bibitem[\protect\citeauthoryear{Quinlan}{Quinlan}{1993}]%
        {Quinlan:1993:CPM:583200}
\bibfield{author}{\bibinfo{person}{J.~Ross Quinlan}.}
  \bibinfo{year}{1993}\natexlab{}.
\newblock \bibinfo{booktitle}{\emph{C4.5: Programs for Machine Learning}}.
\newblock \bibinfo{publisher}{Morgan Kaufmann Publishers Inc.},
  \bibinfo{address}{San Francisco, CA, USA}.
\newblock
\showISBNx{1558602402}


\bibitem[\protect\citeauthoryear{Ribeiro, Singh, and Guestrin}{Ribeiro
  et~al\mbox{.}}{2016}]%
        {ribeiro2016should}
\bibfield{author}{\bibinfo{person}{Marco~Tulio Ribeiro},
  \bibinfo{person}{Sameer Singh}, {and} \bibinfo{person}{Carlos Guestrin}.}
  \bibinfo{year}{2016}\natexlab{}.
\newblock \showarticletitle{Why should i trust you?: Explaining the predictions
  of any classifier}. In \bibinfo{booktitle}{\emph{SIGKDD}}. ACM,
  \bibinfo{pages}{1135--1144}.
\newblock


\bibitem[\protect\citeauthoryear{Rozemberczki, Allen, and Sarkar}{Rozemberczki
  et~al\mbox{.}}{2019}]%
        {rozemberczki2019multiscale}
\bibfield{author}{\bibinfo{person}{Benedek Rozemberczki}, \bibinfo{person}{Carl
  Allen}, {and} \bibinfo{person}{Rik Sarkar}.} \bibinfo{year}{2019}\natexlab{}.
\newblock \bibinfo{title}{Multi-scale Attributed Node Embedding}.
\newblock
\newblock
\showeprint[arxiv]{1909.13021}~[cs.LG]


\bibitem[\protect\citeauthoryear{Scholkopf and Smola}{Scholkopf and
  Smola}{2001}]%
        {scholkopf2001learning}
\bibfield{author}{\bibinfo{person}{Bernhard Scholkopf} {and}
  \bibinfo{person}{Alexander~J Smola}.} \bibinfo{year}{2001}\natexlab{}.
\newblock \bibinfo{booktitle}{\emph{Learning with kernels: support vector
  machines, regularization, optimization, and beyond}}.
\newblock \bibinfo{publisher}{MIT press}.
\newblock


\bibitem[\protect\citeauthoryear{Wang, Rudin, Doshi-Velez, Liu, Klampfl, and
  MacNeille}{Wang et~al\mbox{.}}{2017}]%
        {wang2017bayesian}
\bibfield{author}{\bibinfo{person}{Tong Wang}, \bibinfo{person}{Cynthia Rudin},
  \bibinfo{person}{Finale Doshi-Velez}, \bibinfo{person}{Yimin Liu},
  \bibinfo{person}{Erica Klampfl}, {and} \bibinfo{person}{Perry MacNeille}.}
  \bibinfo{year}{2017}\natexlab{}.
\newblock \showarticletitle{A bayesian framework for learning rule sets for
  interpretable classification}.
\newblock \bibinfo{journal}{\emph{JMLR}} \bibinfo{volume}{18},
  \bibinfo{number}{1} (\bibinfo{year}{2017}), \bibinfo{pages}{2357--2393}.
\newblock


\bibitem[\protect\citeauthoryear{Wang, Zhang, Liu, and Wang}{Wang
  et~al\mbox{.}}{2020}]%
        {wang2020transparent}
\bibfield{author}{\bibinfo{person}{Zhuo Wang}, \bibinfo{person}{Wei Zhang},
  \bibinfo{person}{Ning Liu}, {and} \bibinfo{person}{Jianyong Wang}.}
  \bibinfo{year}{2020}\natexlab{}.
\newblock \showarticletitle{Transparent Classification with Multilayer Logical
  Perceptrons and Random Binarization.}. In \bibinfo{booktitle}{\emph{AAAI}}.
  \bibinfo{pages}{6331--6339}.
\newblock


\bibitem[\protect\citeauthoryear{Xiao, Rasul, and Vollgraf}{Xiao
  et~al\mbox{.}}{2017}]%
        {xiao2017:online}
\bibfield{author}{\bibinfo{person}{Han Xiao}, \bibinfo{person}{Kashif Rasul},
  {and} \bibinfo{person}{Roland Vollgraf}.} \bibinfo{year}{2017}\natexlab{}.
\newblock \bibinfo{title}{Fashion-MNIST: a Novel Image Dataset for Benchmarking
  Machine Learning Algorithms}.
\newblock
\newblock
\showeprint[arXiv]{cs.LG/1708.07747}~[cs.LG]


\bibitem[\protect\citeauthoryear{Yang, Rudin, and Seltzer}{Yang
  et~al\mbox{.}}{2017}]%
        {yang2017scalable}
\bibfield{author}{\bibinfo{person}{Hongyu Yang}, \bibinfo{person}{Cynthia
  Rudin}, {and} \bibinfo{person}{Margo Seltzer}.}
  \bibinfo{year}{2017}\natexlab{}.
\newblock \showarticletitle{Scalable Bayesian rule lists}. In
  \bibinfo{booktitle}{\emph{ICML}}. JMLR. org, \bibinfo{pages}{3921--3930}.
\newblock


\bibitem[\protect\citeauthoryear{Yang, Morillo, and Hospedales}{Yang
  et~al\mbox{.}}{2018}]%
        {yang2018deep}
\bibfield{author}{\bibinfo{person}{Yongxin Yang}, \bibinfo{person}{Irene~Garcia
  Morillo}, {and} \bibinfo{person}{Timothy~M Hospedales}.}
  \bibinfo{year}{2018}\natexlab{}.
\newblock \showarticletitle{Deep neural decision trees}.
\newblock \bibinfo{journal}{\emph{arXiv preprint arXiv:1806.06988}}
  (\bibinfo{year}{2018}).
\newblock


\end{thebibliography}





\section*{Checklist}


\begin{enumerate}

\item For all authors...
\begin{enumerate}
  \item Do the main claims made in the abstract and introduction accurately reflect the paper's contributions and scope?
    \answerYes{}
  \item Did you describe the limitations of your work?
    \answerYes{See Section \ref{section:conclusion}.}
  \item Did you discuss any potential negative societal impacts of your work?
    \answerYes{There is no potential negative societal impact of our work.}
  \item Have you read the ethics review guidelines and ensured that your paper conforms to them?
    \answerYes{}
\end{enumerate}

\item If you are including theoretical results...
\begin{enumerate}
  \item Did you state the full set of assumptions of all theoretical results?
    \answerNA{}
	\item Did you include complete proofs of all theoretical results?
    \answerNA{}
\end{enumerate}

\item If you ran experiments...
\begin{enumerate}
  \item Did you include the code, data, and instructions needed to reproduce the main experimental results (either in the supplemental material or as a URL)?
    \answerYes{See Appendix \ref{appendix:code_and_data} for the code, data, and instructions.}
  \item Did you specify all the training details (e.g., data splits, hyperparameters, how they were chosen)?
    \answerYes{See Section \ref{section:dataset_desc_and_exp_settings} and Appendix \ref{appendix:experimental_setting}.}
	\item Did you report error bars (e.g., with respect to the random seed after running experiments multiple times)?
    \answerYes{See Figure \ref{fig:model_complexity} and Appendix \ref{appendix:model_complexity}.}
	\item Did you include the total amount of compute and the type of resources used (e.g., type of GPUs, internal cluster, or cloud provider)?
    \answerYes{See Appendix \ref{appendix:experimental_setting}.}
\end{enumerate}

\item If you are using existing assets (e.g., code, data, models) or curating/releasing new assets...
\begin{enumerate}
  \item If your work uses existing assets, did you cite the creators?
    \answerYes{}
  \item Did you mention the license of the assets?
    \answerYes{}
  \item Did you include any new assets either in the supplemental material or as a URL?
    \answerYes{}
  \item Did you discuss whether and how consent was obtained from people whose data you're using/curating?
    \answerNA{}
  \item Did you discuss whether the data you are using/curating contains personally identifiable information or offensive content?
    \answerYes{The data we are using do not contain any personally identifiable information or offensive content.}
\end{enumerate}

\item If you used crowdsourcing or conducted research with human subjects...
\begin{enumerate}
  \item Did you include the full text of instructions given to participants and screenshots, if applicable?
    \answerNA{}
  \item Did you describe any potential participant risks, with links to Institutional Review Board (IRB) approvals, if applicable?
    \answerNA{}
  \item Did you include the estimated hourly wage paid to participants and the total amount spent on participant compensation?
    \answerNA{}
\end{enumerate}

\end{enumerate}

\newpage
\appendix



\section{Code Release and Data Source}
\label{appendix:code_and_data}
Our code is publicly available at a GitHub repository: \url{https://github.com/12wang3/rrl}.

The datasets used in this paper come from the UCI machine learning repository and GitHub. The links to the datasets are: \textit{adult}\footnote{\url{https://archive.ics.uci.edu/ml/datasets/Adult}}, \textit{bank-marketing}\footnote{\url{http://archive.ics.uci.edu/ml/datasets/Bank+Marketing}}, \textit{banknote}\footnote{\url{https://archive.ics.uci.edu/ml/datasets/banknote+authentication}}, \textit{chess}\footnote{\url{https://archive.ics.uci.edu/ml/datasets/Chess+\%28King-Rook+vs.+King\%29}}, \textit{connect-4}\footnote{\url{http://archive.ics.uci.edu/ml/datasets/connect-4}}, \textit{letRecog}\footnote{\url{https://archive.ics.uci.edu/ml/datasets/Letter+Recognition}}, \textit{magic04}\footnote{\url{https://archive.ics.uci.edu/ml/datasets/magic+gamma+telescope}}, \textit{tic-tac-toe}\footnote{\url{https://archive.ics.uci.edu/ml/datasets/Tic-Tac-Toe+Endgame}}, \textit{wine}\footnote{\url{https://archive.ics.uci.edu/ml/datasets/wine}}, \textit{activity}\footnote{\url{https://archive.ics.uci.edu/ml/datasets/human+activity+recognition+using+smartphones}}, \textit{dota2}\footnote{\url{https://archive.ics.uci.edu/ml/datasets/Dota2+Games+Results}}, \textit{facebook}\footnote{\url{https://archive.ics.uci.edu/ml/datasets/Facebook+Large+Page-Page+Network}}, \textit{fashion}\footnote{\url{https://github.com/zalandoresearch/fashion-mnist}}. The \textit{fashion} dataset uses the MIT license, and other datasets have no license mentioned. The citation requests of these datasets are all satisfied in our paper.

\section{Data Sets Properties}
\label{appendix:data_propertity}
In Table \ref{tab:appendix_property}, the first nine data sets are small data sets while the last four are large data sets. Discrete or continuous feature type indicates features in that data set are all discrete or all continuous. The mixed feature type indicates the corresponding data set has both discrete and continuous features. The density is the averaged ratio of the number of 1 in each binary feature vector after one-hot encoding.

\begin{minipage}{\textwidth}
\centering
	\captionof{table}{Data sets properties.}
	\label{tab:appendix_property}
	\begin{tabular}{cccccc}
\toprule
Dataset & \#instances & \#classes	& \#features & feature type & density\\
\midrule
adult	&	32561	&	2	&	14	&	mixed	&	-	\\
bank-marketing	&	45211	&	2	&	16	&	mixed	&	-	\\
banknote	&	1372	&	2	&	4	&	continuous	&	-	\\
chess	&	28056	&	18	&	6	&	discrete	&	0.150	\\
connect-4	&	67557	&	3	&	42	&	discrete	&	0.333	\\
letRecog	&	20000	&	26	&	16	&	continuous	&	-	\\
magic04	&	19020	&	2	&	10	&	continuous	&	-	\\
tic-tac-toe	&	958	&	2	&	9	&	discrete	&	0.333	\\
wine	&	178	&	3	&	13	&	continuous	&	-	\\
\midrule											
activity	&	10299	&	6	&	561	&	continuous	&	-	\\
dota2	&	102944	&	2	&	116	&	discrete	&	0.087	\\
facebook	&	22470	&	4	&	4714	&	discrete	&	0.003	\\
fashion	&	70000	&	4	&	784	&	continuous	&	-	\\
      \bottomrule
    \end{tabular}
\end{minipage}

\section{Experimental Setting}
\label{appendix:experimental_setting}
\noindent \textbf{Experiment Environment.} We implement our model with PyTorch \citep{paszke2019pytorch}, an open-source machine learning framework. Experiments are conducted on a Linux server with an Intel Xeon E5 v4 CPU at 2.10GHz and one GeForce RTX 2080 Ti GPU.

\noindent \textbf{Parameter Settings.} The number of logical layers in RRL ranges from 1 to 3. The number of nodes in logical layers ranges from 16 to 4096 depending on the number of binary features of the data set and the model complexity we need. 
We use the cross-entropy loss during the training. The L2 regularization is adopted to restrict the model complexity, and the coefficient of the regularization term in the loss function is in $\{10^{-2}, 10^{-3},\dots, 10^{-8}, 0\}$.
The numbers of the lower and upper bounds in the binarization layer are both in $\{5, 10, 50\}$. We utilize the Adam \citep{kingma2014adam} method for the training process with a mini-batch size of 32. The initial learning rate is in $\{5\times 10^{-3},2\times 10^{-3},1\times 10^{-3},5\times 10^{-4},2\times 10^{-4},1\times 10^{-4}\}$.
On small data sets, RRL is trained for 400 epochs, and we decay the learning rate by a factor of 0.75 every 100 epochs.
On large data sets, RRL is trained for 100 epochs, and we decay the learning rate by a factor of 0.75 every 20 epochs. We use the  derivative estimation trick mentioned in Section \ref{section:logical_layer} for RRL trained on large data sets. When parameter tuning is required, 95\% of the training set is used for training and 5\% for validation.

We use validation sets to tune hyperparameters of all the baselines mentioned in Section \ref{section:classification_performance}. We use sklearn to implement LR, and use the L1 or L2 norm in the penalization. The liblinear is used as the solver. The tolerance for stopping criteria is in \{$10^{-3}$,  $10^{-4}$, $10^{-5}$\}. The inverse of regularization strength is in \{1, 4, 16, 32\}. 
For decision tree, its max depth is in \{None, 5, 10, 20\}. The min number of samples required to split an internal node is in \{2, 8, 16\}, and the min number of samples required to be at a leaf node is in \{1, 8, 16\}. 
For SBRL, its $\lambda$ is set to 5 initially, and the min and max rule sizes are set at 1 and 3, respectively. $\eta$ is set to 1, and the numbers of iterations and chains are set to 5000 and 20, respectively. The  minsupport\_pos and minsupport\_neg are set to keep the total number of rules  close to 300.
For CORELS, the regularization parameter is in \{0, $10^{-2}$, $10^{-3}$,  $10^{-4}$, $10^{-5}$\}, and the maximum number of rule lists to search is in \{$10^{4}$, $10^{5}$,  $10^{6}$, $10^{7}$\}. The maximum cardinality allowed is set to 2 or 3, and the min support rate is in \{0.005, 0.01, 0.025, 0.05\}.
For CRS and PLNN, the candidate sets of learning rate, learning rate decay rate, batch size, model structure (depth and width), and coefficient of the regularization term in the loss function are the same as RRL's. The random binarization rate of CRS is in \{0, 0.7, 0.75, 0.8, 0.85, 0.9, 0.95\}.
For SVM, the linear, RBF and Ploy kernels are used. The tolerance for stopping criteria is in \{$10^{-3}$,  $10^{-4}$, $10^{-5}$\}. The inverse of regularization strength is in \{1, 4, 16, 32\}. The kernel coefficient is set to the reciprocal of the number of features.
For Random Forest, its min number of samples required to split an internal node and the min number of samples required to be at a leaf node are the same as the decision tree's. We use LightGBM and XGBoost to implement Gradient Boosted Decision Tree (GBDT). The learning rate of GBDT is in \{0.1, 0.01, 0.001\}, and the max depth of one tree is in \{None, 5, 10, 20\}. The number of estimators in ensemble models is in \{10, 100, 500\}.
For baselines that can not directly deal with the continuous value inputs, we use the recursive minimal entropy partitioning algorithm or the KBinsDiscretizer implemented by sklearn to discretize the inputs first.
Grid search is also used for the parameter tuning.

\section{Vanishing Gradient Problem}
\label{vanishing_gradient_problem}
The partial derivative of each node in $\hat{\mathcal{S}}^{(l)}$ w.r.t. its directly connected weights and w.r.t. its directly connected nodes are given by:
\begin{equation}
\label{eq:derivative_sw}
\frac{\partial \hat{\mathbf{s}}_{i}^{(l)}}{\partial \hat{W}_{i,j}^{(l,1)}}
=\hat{\mathbf{u}}_{j}^{(l-1)} \cdot \prod_{k \neq j}(1-F_{d}(\hat{\mathbf{u}}_{k}^{(l-1)}, \hat{W}_{i,k}^{(l,1)}))
\end{equation}
\begin{equation}
\label{eq:derivative_su}
\frac{\partial \hat{\mathbf{s}}_{i}^{(l)}}{\partial \hat{\mathbf{u}}_{j}^{(l-1)}}
=\hat{W}_{i,j}^{(l,1)} \cdot \prod_{k \neq j}(1-F_{d}(\hat{\mathbf{u}}_{k}^{(l-1)}, \hat{W}_{i,k}^{(l,1)}))
\end{equation}
Similar to the analysis of Equation \ref{eq:derivative_rw_ru}, due to $\hat{\mathbf{u}}_{k}^{(l-1)}$ and $\hat{W}_{i,k}^{(l,1)}$ are in the range $[0, 1]$, the values of $(1-F_d(\cdot))$ in Equation \ref{eq:derivative_sw} and \ref{eq:derivative_su} are in the range $[0, 1]$ as well. If $\mathbf{n}_{l-1}$ is large and most of the values of $(1-F_d(\cdot))$ are not 1, then the derivative is close to 0 because of the multiplications. Therefore, both the conjunction function and the disjunction function suffer from vanishing gradient problem.

If the large data set is very sparse and the number of 1 in each binary feature vector (for RRL the binary feature vector is $\mathbf{u}^{(0)}$) is less than about one hundred, there will be no vanishing gradient problem for nodes in $\hat{\mathcal{S}}^{(1)}$. The reason is when the number of 1 in each feature vector is less than about one hundred, in Equation \ref{eq:derivative_sw} and \ref{eq:derivative_su}, most of the values of $(1-F_d(\cdot))$ are 1, and only less than one hundred values of $(1-F_d(\cdot))$ are not 1, then the result of the multiplication is not very close to 0. The \textit{facebook} data set is an example of this case. However, if the number of 1 in each binary feature vector is more than about one hundred, the vanishing gradient problem comes again.
\section{Gradients at Discrete Points}
\label{gradients_at_discrete_points}
The gradients of RRL with original logical activation functions at discrete points can be obtained by Equation \ref{eq:derivative_rw_ru}, \ref{eq:derivative_sw} and \ref{eq:derivative_su}. Take Equation \ref{eq:derivative_sw} as an example, discrete points mean all the weights of logical layers are 0 or 1, which also means the values of all the nodes in $\hat{\mathcal{U}}^{(l)}$ are 0 or 1, $l\in\{0,1,\dots,L-1\}$. Hence, in Equation \ref{eq:derivative_sw}, $\hat{\mathbf{u}}_{j}^{(l-1)}, (1-F_d(\cdot)) \in \{0,1\}$, and the whole equation is actually multiplications of several 0 and several 1. Only when $\hat{\mathbf{u}}_{j}^{(l-1)}$ and $(1-F_d(\cdot))$ are all 1, the derivative in Equation \ref{eq:derivative_sw} is 1, otherwise, the derivative is 0. Therefore, the gradients at discrete points have no useful information in most cases. The analyses of Equation \ref{eq:derivative_rw_ru} and \ref{eq:derivative_su} are similar.

To analyze the gradients of RRL with improved logical activation functions at discrete points, we first calculate the partial derivative of each node w.r.t. its directly connected weights and w.r.t. its directly connected nodes:
\begin{equation}
\label{eq:derivative_irw}
\frac{\partial \hat{\mathbf{r}}_{i}^{(l)}}{\partial \hat{W}_{i,j}^{(l,0)}}
=\frac{(\hat{\mathbf{r}}_{i}^{(l)})^2 }{F_c(\hat{\mathbf{u}}^{(l-1)}_j, \hat{W}^{(l,0)}_{i,j})+\epsilon}\cdot (\hat{\mathbf{u}}^{(l-1)}_j-1)
\end{equation}
\begin{equation}
\label{eq:derivative_iru}
\frac{\partial \hat{\mathbf{r}}_{i}^{(l)}}{\partial \hat{\mathbf{u}}^{(l-1)}_j}
=\frac{(\hat{\mathbf{r}}_{i}^{(l)})^2 }{F_c(\hat{\mathbf{u}}^{(l-1)}_j, \hat{W}^{(l,0)}_{i,j})+\epsilon}\cdot \hat{W}^{(l,0)}_{i,j}
\end{equation}

\begin{equation}
\label{eq:derivative_isw}
\frac{\partial \hat{\mathbf{s}}_{i}^{(l)}}{\partial \hat{W}_{i,j}^{(l,1)}}
=\frac{(1-\hat{\mathbf{s}}_{i}^{(l)})^2}{1-F_d(\hat{\mathbf{u}}_{j}^{(l-1)},\hat{W}_{i,j}^{(l,1)})+\epsilon}\cdot \hat{\mathbf{u}}^{(l-1)}_j
\end{equation}

\begin{equation}
\label{eq:derivative_isu}
\frac{\partial \hat{\mathbf{s}}_{i}^{(l)}}{\partial \hat{\mathbf{u}}_{j}^{(l-1)}}
=\frac{(1-\hat{\mathbf{s}}_{i}^{(l)})^2}{1-F_d(\hat{\mathbf{u}}_{j}^{(l-1)},\hat{W}_{i,j}^{(l,1)})+\epsilon}\cdot \hat{W}_{i,j}^{(l,1)}
\end{equation}
Take Equation \ref{eq:derivative_irw} for example, when all the weights of logical layers are 0 or 1, the $\hat{\mathbf{r}}_{i}^{(l)}$, $F_c(\cdot)+\epsilon$ and $(\hat{\mathbf{u}}^{(l-1)}_j-1)$ are all very close to 0 or 1 as well. For the initialized weights are randomly selected, $\hat{\mathbf{r}}_{i}^{(l)}$ is close to 0 in most cases. Hence, the derivative in Equation \ref{eq:derivative_irw} is close to 0 in most cases, and the analyses of Equation \ref{eq:derivative_iru}, \ref{eq:derivative_isw} and \ref{eq:derivative_isu} are similar. Therefore, the gradients at discrete points have little useful information. 

\section{Computation Time}
The computation time of RRL is similar to neural networks like Multilayer Perceptrons (MLP) for their computations are quite similar. The training time of RRL (in Table \ref{tab:accuracy}) on all the datasets (400 epochs on the small datasets and 100 epochs on the large datasets) with one GeForce RTX 2080 Ti is shown in Table \ref{tab:appendix_time}. We can see that the training time of RRL is acceptable on all the datasets, which also verifies the good scalability of RRL.

\noindent \begin{minipage}{1.\textwidth}
\vspace{\baselineskip}
\centering
	\captionof{table}{Training time of RRL on nine small and four large datasets.}
	\label{tab:appendix_time}
	\resizebox{0.9\linewidth}{!}{
	\begin{tabular}{cccccccc}
\toprule
Dataset &	adult	&	bank-marketing	&	banknote	&	chess	&	connect-4	&	letRecog	\\
\midrule
Time	&	1h22m55s	&	1h3m41s	&	7m45s	&	29m40s	&	2h20m41s	&	2h16m24s\\
      \bottomrule
    \end{tabular}
    }
\end{minipage}
\begin{minipage}{1.\textwidth}
\centering
	\resizebox{0.9\linewidth}{!}{
	\begin{tabular}{cccccccp{30pt}}
\toprule
Dataset	&	magic04 &	tic-tac-toe	&	wine	&	activity	&	dota	&	facebook	&	fashion	   \\
\midrule
Time	&	3h22m37s	&	1m12s	&	16s	&	1h2m24s	&	1h58m42s	&	2h27m23s	&	7h32m52s	  \\
      \bottomrule
    \end{tabular}
    }
\end{minipage}
\section{Average Rule Length}
The average length of rules in RRL (in Table \ref{tab:accuracy}) trained on different datasets is shown in Table \ref{tab:appendix_avglen}. We can observe that, except for the \textit{facebook} and \textit{fashion} datasets, the average length of rules is less than 13 (most are less than 7), which means understanding one rule is easy and understanding the rules one by one in the order of weights is feasible. The average length of rules of RRL trained on the \textit{facebook} and \textit{fashion} datasets is large for \textit{facebook} and \textit{fashion} are actually two unstructured datasets, e.g., the \textit{fashion} dataset is an image classification dataset. 

\noindent \begin{minipage}{1.\textwidth}
\vspace{\baselineskip}
\centering
	\captionof{table}{Average rule length of RRL trained on nine small and four large datasets.}
	\label{tab:appendix_avglen}
	\resizebox{0.9\linewidth}{!}{
	\begin{tabular}{cccccccc}
\toprule
Dataset &	adult	&	bank-marketing	&	banknote	&	chess	&	connect-4	&	letRecog\\
\midrule
AvgLength		&	5.71	&	3.78	&	3.56	&	7.03	&	12.44	&	5.91	\\
      \bottomrule
    \end{tabular}
    }
\end{minipage}
\begin{minipage}{1.\textwidth}
\centering
	\resizebox{0.9\linewidth}{!}{
	\begin{tabular}{cccccccp{30pt}}
\toprule
Dataset	&	magic04	 &	tic-tac-toe	&	wine	&	activity	&	dota	&	facebook	&	fashion	  \\
\midrule
AvgLength	&	5.05	&	2.88	&	2.11	&	6.67	&	4.37	&	38.28	&	34.53     \\
      \bottomrule
    \end{tabular}
    }
\end{minipage}

\section{Classification Performance}
\label{appendix:classification_performance}

\begin{minipage}{\textwidth}
\centering
	\captionof{table}{5-fold cross validated accuracy (\%) of comparing models on all 13 datasets. $^*$ represents that RRL significantly outperforms all the compared interpretable models (t-test with p < 0.01). The first seven models are interpretable models, while the last five are complex models.}
	\label{tab:appendix_accuracy}
\resizebox{1.\linewidth}{!}{
  \begin{tabular}{c|ccccccc|ccccc}
    \toprule
    Dataset & \textbf{RRL} & C4.5 & CART & SBRL & CORELS & CRS & LR & SVM & PLNN(MLP) & RF & LightGBM & XGBoost\\
\midrule
adult & 85.73 & 85.28 & 85.19 & 84.83 & 79.98 & 85.71 & 85.25 & 79.78 & 85.71 & 85.57 & 85.96 & \textbf{86.13}\\
bank-marketing & 90.63 & 89.93 & 89.83 & 90.38 & 88.30 & 90.11 & 90.13 & 88.31 & 90.54 & 90.68 & 90.76 & \textbf{90.91}\\
banknote & \textbf{100.00}$^{*}$ & 98.61 & 98.03 & 93.95 & 97.08 & 93.46 & 98.83 & \textbf{100.00} & \textbf{100.00} & 99.56 & 99.64 & 99.56\\
chess & 82.02 & 79.20 & 78.67 & 29.37 & 32.39 & 82.38 & 36.07 & 82.17 & 77.05 & 75.14 & 84.93 & \textbf{85.40}\\
connect-4 & \textbf{86.96}$^{*}$ & 76.32 & 76.38 & 62.80 & 64.93 & 79.05 & 75.74 & 83.65 & 85.39 & 83.10 & 85.53 & 86.47\\
letRecog & 95.59$^{*}$ & 88.18 & 87.51 & 62.88 & 58.48 & 83.13 & 72.31 & 94.92 & 92.48 & 96.54 & \textbf{96.86} & 96.45\\
magic04 & 87.52$^{*}$ & 84.77 & 83.72 & 83.58 & 79.09 & 83.81 & 79.02 & 82.68 & 86.49 & 88.20 & 88.46 & \textbf{88.62}\\
tic-tac-toe & \textbf{100.00} & 94.26 & 93.32 & 98.85 & 98.75 & 99.06 & 98.33 & 98.33 & 98.12 & 98.75 & 99.90 & 99.27\\
wine & 97.76 & 93.83 & 90.46 & 92.70 & 94.52 & 97.22 & 95.54 & 91.03 & 90.52 & \textbf{97.78} & 96.08 & 96.06\\
\midrule
activity & 97.96 & 94.08 & 93.31 & 18.25 & 61.04 & 17.33 & 98.33 & 98.29 & 98.28 & 97.92 & \textbf{99.37} & 99.26\\
dota2 & \textbf{59.87} & 52.92 & 52.73 & 51.91 & 53.02 & 55.95 & 59.46 & 59.57 & 55.27 & 58.20 & 58.80 & 59.35\\
facebook & \textbf{90.53}$^{*}$ & 82.59 & 82.90 & 49.72 & 41.27 & 28.91 & 89.44 & 88.46 & 90.00 & 88.32 & 86.85 & 88.00\\
fashion & 89.22$^{*}$ & 80.68 & 79.43 & 54.57 & 49.14 & 67.71 & 84.78 & 89.34 & 89.29 & 88.53 & \textbf{91.35} & 90.71\\
\midrule
\textbf{AvgRank} & 2.92 & 8.15 & 9.23 & 9.92 & 10.23 & 7.62 & 7.23 & 6.15 & 5.62 & 5.08 & \textbf{2.77} & \textbf{2.77}\\
  \bottomrule
\end{tabular}
}
\end{minipage}

\section{Case Study}
\label{appendix:case_study}
Although RRL is not designed for image classification tasks, due to its high scalability, it can still provide intuition by visualizations. Take the \textit{fashion} dataset for example, for each class, we combine the first ten rules, ordered by linear layer weights, for feature (pixel) visualization. In Figure \ref{fig:fashion}, a black/white pixel indicates the combined rule asks for a color close to black/white here in the original input image, and the grey pixel means no requirement in the rule. According to these figures, we can see how RRL classifies the images, e.g., distinguishing T-shirt from Pullover by sleeves.

\begin{minipage}{\textwidth}
\centering
    \includegraphics[width=0.8\textwidth]{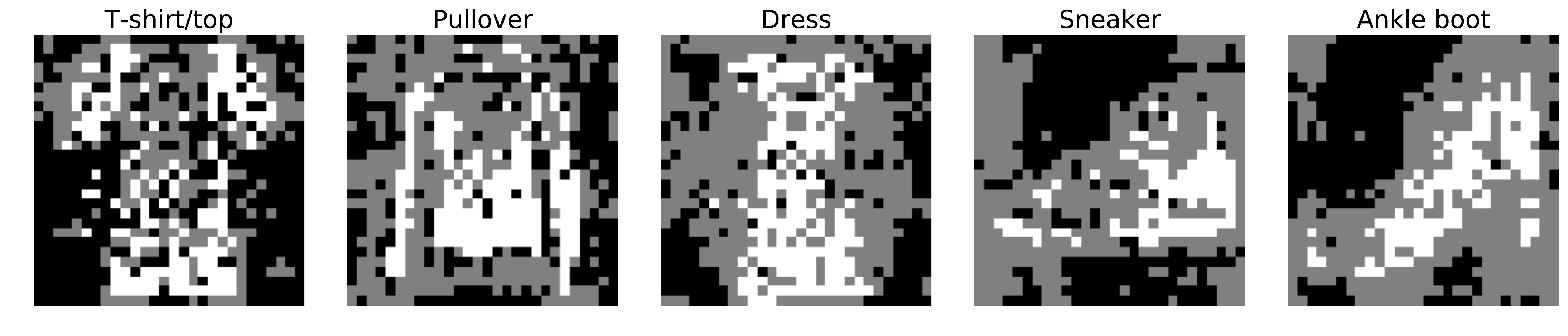}
    \captionof{figure}{Decision mode for the \textit{fashion} data set summarized from rules of RRL.}
    \label{fig:fashion}
\end{minipage}

\section{Model Complexity}
\label{appendix:model_complexity}
\begin{figure}
    \centering
    \includegraphics[width=0.66\textwidth]{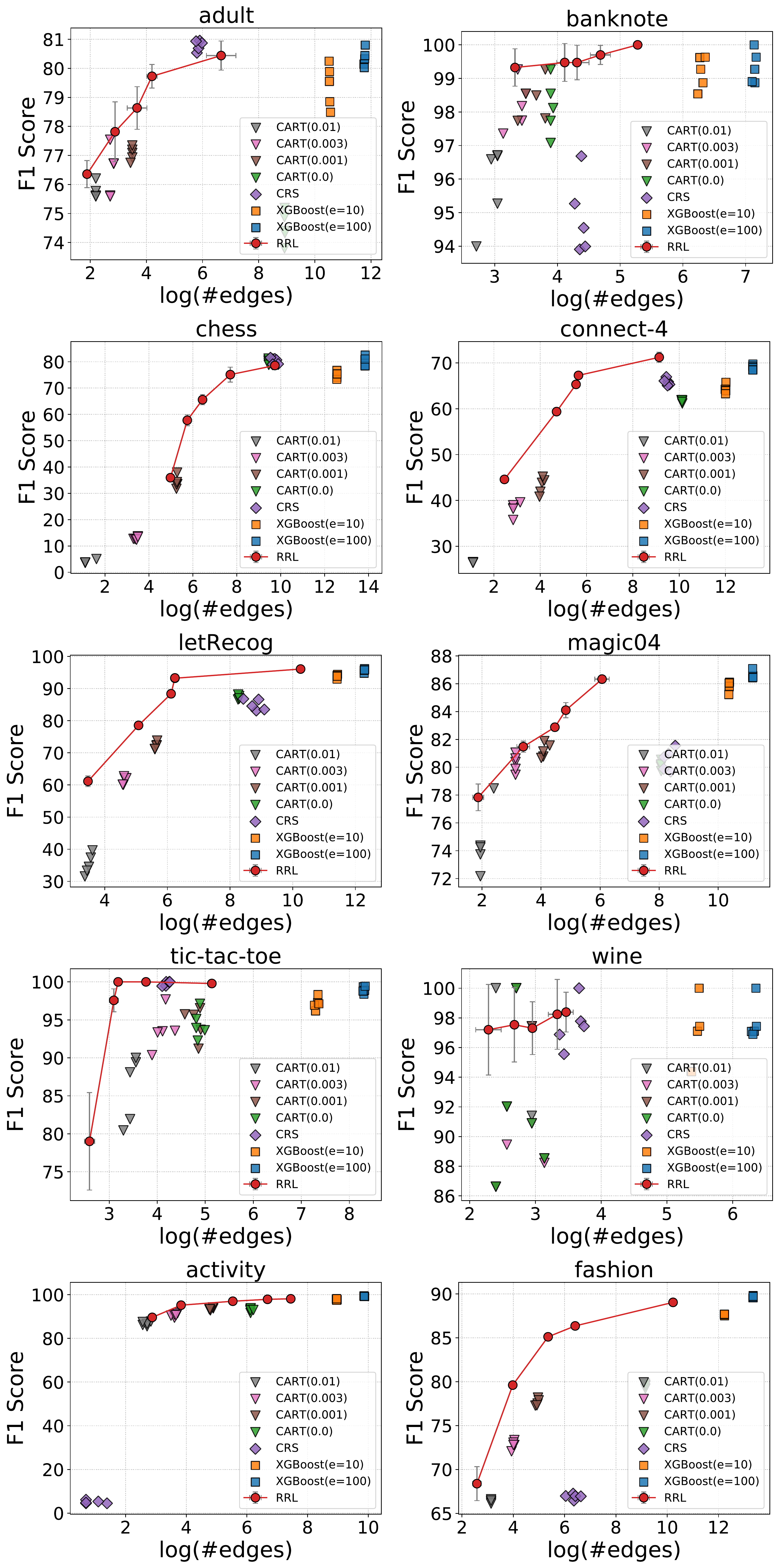}
    \caption{Scatter plot of F1 score against log(\#edges) for RRL and baselines on ten datasets.}
    \label{fig:appendix_complexity_appendix}
\end{figure}

Figure \ref{fig:appendix_complexity_appendix} shows the scatter plots of F1 score against log(\#edges) for rule-based models trained on the other ten data sets. For RRL, the legend markers and error bars indicate means and standard deviations, respectively, of F1 score and log(\#edges) across cross-validation folds.
For baseline models, each point represents an evaluation of one model, on one fold, with one parameter setting.
The value in CART($\cdot$), e.g., CART(0.03), denotes the complexity parameter used for Minimal Cost-Complexity Pruning \citep{breiman2017classification}, and a higher value corresponds to a simpler tree. We also show the results of XGBoost with 10 and 100 estimators.
On these ten data sets, we can still observe that if we connect the results of RRL, it will become a boundary that separating the upper left corner from other models. In other words, if RRL has a close model complexity with one baseline, then the F1 score of RRL will be higher, or if the F1 score of RRL is close to one baseline, then its model complexity will be lower. It indicates that RRL can make better use of rules than rule-based models using heuristic and ensemble methods in most cases.

\end{document}